\documentclass[10pt,twocolumn,letterpaper]{article}
\usepackage{authblk}

\usepackage[pagenumbers]{cvpr} %

\usepackage{epsfig}
\usepackage{graphicx}
\usepackage{comment}
\usepackage{amsmath,amssymb} %
\usepackage{xcolor}
\usepackage[export]{adjustbox}
\usepackage{bbm}

\usepackage[font=small,skip=5pt]{caption}
\usepackage{float}

\newcommand{\specialcell}[2][c]{%
  \begin{tabular}[#1]{@{}c@{}}#2\end{tabular}}
\usepackage{multirow}
\usepackage{tabularx}
\usepackage{soul}
\usepackage{array}
\usepackage{comment}
\usepackage{outlines}

\usepackage[mathscr]{euscript}
\newcolumntype{L}[1]{>{\raggedright\let\newline\\arraybackslash\hspace{0pt}}m{#1}}
\newcolumntype{C}[1]{>{\centering\let\newline\\arraybackslash\hspace{0pt}}m{#1}}
\newcolumntype{R}[1]{>{\raggedleft\let\newline\\arraybackslash\hspace{0pt}}m{#1}}
\newcolumntype{Y}{>{\centering\arraybackslash}X}
\usepackage{booktabs}
\usepackage{enumitem}

\usepackage[pagebackref,breaklinks,bookmarks=false,colorlinks]{hyperref}
\usepackage{amsmath}

\newcommand{\method}{BEHAVE}

\newcommand{\funct}[1]{\boldsymbol{#1}}
\newcommand{\mat}[1]{\mathbf{#1}}
\newcommand{\set}[1]{\mathcal{#1}}
\newcommand{\vect}[1]{\mathbf{#1}}

\newcommand{\pose}[0]{\boldsymbol{\theta}}
\newcommand{\shape}[0]{\boldsymbol{\beta}}

\newcommand{\smpl}[0]{M}

\newcommand{\net}{f_\phi}

\newcommand{\humanpc}{\set{S}^h}
\newcommand{\objectpc}{\set{S}^o}

\newcommand\figONEsize{0.97}
\newcommand\figTWOsize{0.12}
\newcommand\figTHREEsize{0.99}
\newcommand\figFOURsize{0.3}
\newcommand\figFIVEsize{0.303}
\newcommand\figSIXsize{0.194}

\usepackage[capitalize]{cleveref}
\crefname{section}{Sec.}{Secs.}
\Crefname{section}{Section}{Sections}
\Crefname{table}{Table}{Tables}
\crefname{table}{Tab.}{Tabs.}

\def\cvprPaperID{869} %
\def\confName{CVPR}
\def\confYear{2022}

\begin{document}

\title{{BEHAVE}: Dataset and Method for Tracking Human Object Interactions}

\author[1,2]{Bharat Lal Bhatnagar}%
\author[2]{Xianghui Xie}
\author[1]{Ilya A. Petrov}
\author[3]{Cristian Sminchisescu}
\author[2]{Christian Theobalt}
\author[1,2]{Gerard Pons-Moll}
\affil[1]{University of T\"ubingen, Germany}
\affil[2]{Max Planck Institute for Informatics, Saarland Informatics Campus, Germany}
\affil[3]{Google Research}

\affil[ ]{\tt\small \{i.petrov, gerard.pons-moll\}@uni-tuebingen.de,
\{bbhatnag, xxie, theobalt\}@mpi-inf.mpg.de,
sminchisescu@google.com}

\maketitle

\begin{abstract}
Modelling interactions between humans and objects in natural environments is central to many applications including gaming, virtual and mixed reality, as well as human behavior analysis and human-robot collaboration. This challenging operation scenario requires generalization to vast number of objects, scenes, and human actions.
Unfortunately, there exist no such dataset.
Moreover, this data needs to be acquired in diverse natural environments, which rules out 4D scanners and marker based capture systems.
We present \method{} dataset, the first full body human-object interaction dataset with multi-view RGBD frames and corresponding 3D SMPL and object fits along with the annotated contacts between them.
We record  $\sim$15k frames at 5 locations with 8 subjects performing a wide range of interactions with 20 common objects.
We use this data to learn a model that can jointly track humans and objects in natural environments with an easy-to-use portable multi-camera setup.
Our key insight is to predict correspondences from the human and the object to a statistical body model to obtain human-object contacts during interactions.
Our approach can record and track not just the humans and objects but also their interactions, modeled as surface contacts, in 3D.
Our code and data can be found at: \href{http://virtualhumans.mpi-inf.mpg.de/behave}{http://virtualhumans.mpi-inf.mpg.de/behave}.
\end{abstract}

\section{Introduction}
\label{sec:intro}
\begin{figure}[t]
  \centering
  \fbox{\includegraphics[width=\figONEsize\linewidth]{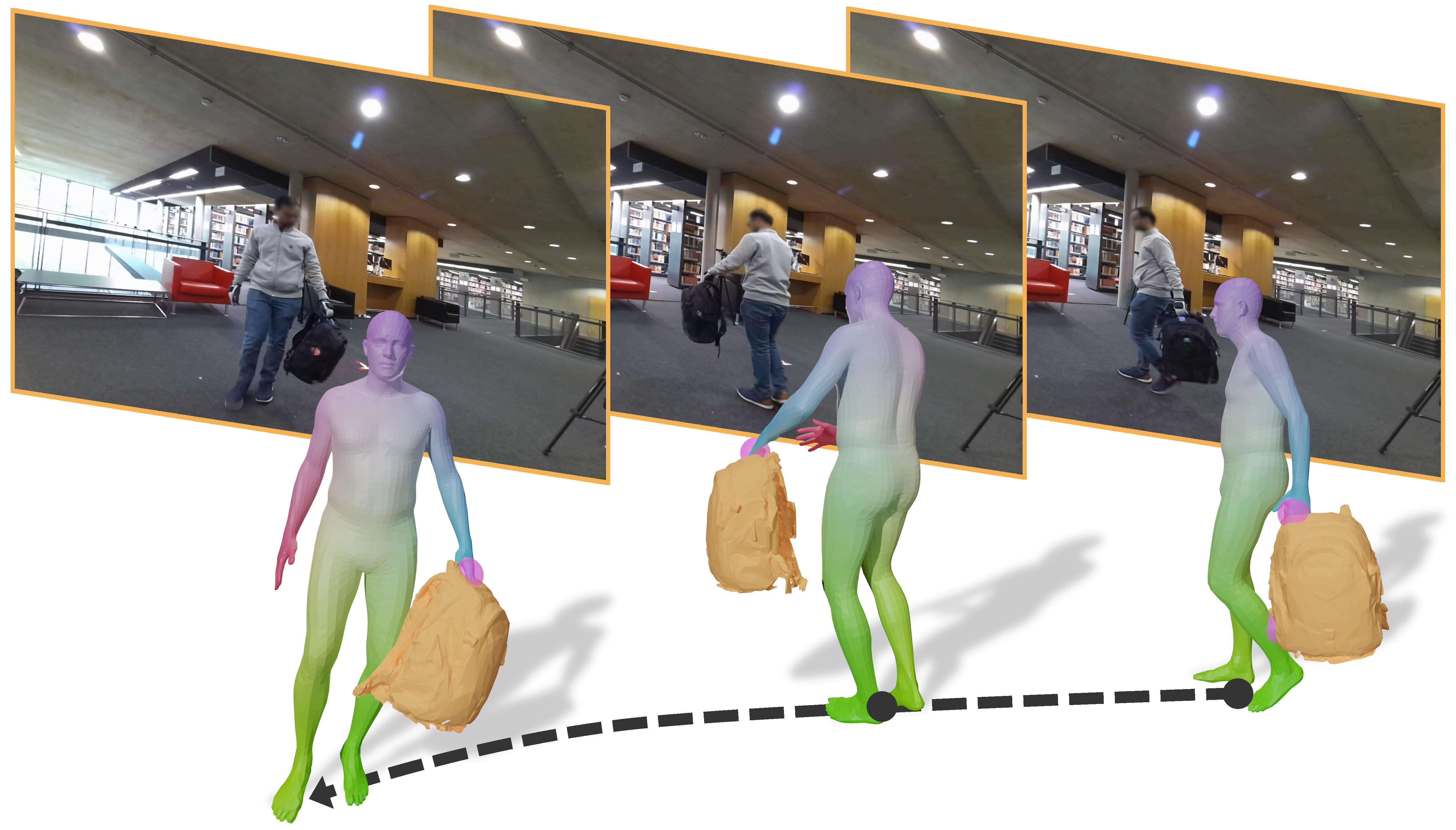}}
  \caption{Given a multi-view RGBD sequence, our method tracks the human, the object and their contacts in 3D.}
  \label{fig:teaser}
\end{figure}

The last decade has seen rapid progress in modelling the appearance of humans ranging from body pose, shape~\cite{smpl2015loper,SMPL-X:2019,xu2020ghum,pons2015dyna,ponsmoll2017clothcap}, faces~\cite{tewari17MoFA} and even detailed clothing~\cite{alldieck2017optical,alldieck2018detailed,bhatnagar2019mgn,mh_patel2020, saito2020pifuhd}.
With various practical use cases like virtual try-on, personalised avatar creation, and several applications in augmented and mixed reality, or human-robot collaboration, the focus on humans is justified.
Beyond modelling appearance, few methods have focused on capturing and synthesizing \emph{human interactions} (human-object/scene interaction). 
There exists work to capture humans in a static 3D scene~\cite{PROX:2019}, even without using external cameras~\cite{HPS}, and work to synthesize static poses~\cite{3d-affordance,Hassan:CVPR:2021}, or full body movement~\cite{nsm,hassan_samp_2021,motionvae,hassan_samp_2021} in a 3D scene.

These methods show growing interest in modelling human behavior, highlighting a need to capture real human interactions.
Existing methods~\cite{nsm,hassan_samp_2021} however are learned from high quality curated data captured using optical marker based motion capture systems or wearable sensors.
Unfortunately, such commercial systems are expensive, drastically limit the interactions that can be captured, and often fail when tracking humans and objects under occlusion.
In addition, the recording volume is spatially confined and difficult to re-locate, thus limiting the activities, scenes, and objects that can be captured. 
Wearable sensors~\cite{HPS} are not restricted in volume, but close range interaction can not be accurately captured.
Altogether, the lack of diverse 3D interaction data, and the lack of accurate and flexible capture methods both constitute barriers in modelling human behavior.

With the goal of simplifying the data capture process and hence allowing faster progress in the field, 
we propose \method{}, a method to capture diverse 3D human interactions in natural environments, using a setup comprising of portable, cheap, and easy to use RGBD cameras.
Tracking human interactions from sparse consumer grade cameras is however extremely challenging. 
Depth data is inherently noisy and incomplete. Moreover, the person and object occlude each other frequently during interactions.
Furthermore, capturing interactions requires estimating human-object contacts accurately, which is difficult because contacts represent small regions in the image, close to the observable (resolution) limit. 
This requires innovation that goes significantly beyond the current state of the art trackers.
We propose to track the human using a parametric human model (such as SMPL~\cite{smpl2015loper}) and track objects using template meshes.
Naively fitting the human model and an object 3D template to the point-cloud completely fails due to the aforementioned challenges. 
Our key idea is to train a neural model which jointly completes the human and object shape, represented with implicit surfaces, 
while predicting a correspondence field to the human, as well as an object orientation field. These rich outputs allow us to formulate a powerful human-object fitting objective which is robust to missing data, noise and occlusion. 

To train and evaluate \method{}, we capture the \emph{largest} dataset of human-object interactions in natural environments. The \method{} dataset contains 20 3D objects, 8 subjects (5 male, 3 female), 5 different locations and totals around 15.2k frames of recording. We provide ground truth SMPL and 3D object meshes as well as contacts.
\\
Our contributions can be summarized as follows:
\begin{itemize}
    \item We propose the first approach that can accurately 3D track humans, objects and contacts in natural environments using multi-view RGBD images.   
    \item We collect the \emph{largest} dataset of multi-view RGBD sequences and corresponding human models, object and contact annotations. See ~\cref{sec:dataset} for details regarding its usefulness to the community.
    \item Since there exists no publicly available code and datasets to accurately track human-object interactions in natural environments, we will release our code and data for further research in this direction.
\end{itemize}

\section{Related Work}
\label{sec:related_work}
In this section, we first briefly review work focused on object and human reconstruction, in isolation from their environmental context. Such methods focus on modelling appearance and do not consider interactions.
Next, we cover methods focused on humans in static scenes and finally discuss closer-related work to ours, for modelling dynamic human-object interactions.

\subsection{Appearance modelling: Humans and objects without scene context}

\paragraph{Human reconstruction and performance capture}
Perceiving humans from monocular RGB data \cite{bogo2016smplify,  kanazawa2018endtoend, pavlakos2018humanshape, Guler_2019_CVPR, kolotouros2019spin, pifuSHNMKL19, SMPL-X:2019, VIBE:CVPR:2020, habermann2020deepcap,Zanfir_2021_ICCV} and under multiple views \cite{huang2017towards, joo2018total, rhodin2018learning, iskakov2019learnable, huang2018deep} settings has been widely explored.  Recent work tends to focus on reconstructing fine details like hand gestures and facial expressions \cite{choutas2020monocular, zhou2021monocular, zanfir2020neural,PIXIE:3DV:2021}, self-contacts \cite{fieraru2021learning,muller2021self}, interactions between humans\cite{fieraru2020three}, and even clothing~\cite{alldieck2019learning,bhatnagar2019mgn}. These methods benefit from representing human with parametric body models~\cite{smpl2015loper,SMPL-X:2019,xu2020ghum}, thus motivating our use of recent implicit diffused representations~\cite{bhatnagar2020loopreg,alldieck2020imghum} as backbone for our tracker.
\\
Following the success of pixel-aligned implicit function learning \cite{pifuSHNMKL19, saito2020pifuhd}, recent methods can capture human performance from sparse~\cite{huang2018deep, xu2021hnerf} or even a single RGB camera~\cite{li-monocap, li2020monocular}.  However, capturing 3D humans from RGB data involves a fundamental ambiguity between depth and scale.
Therefore, recent methods use RGBD~\cite{DoubleFusion2018, pandey2019volumetric, wang2020normalgan, su2020robustfusion, tao2021function4d} or volumetric data~\cite{chibane2020implicit, bhatnagar2020loopreg, bhatnagar2020ipnet} for reliable human capture. These insights motivate us to build novel trackers based on multi-view RGBD data.

\paragraph{Object reconstruction}
Most existing work on reconstructing 3D objects from RGB~\cite{tzionas20153d, choy20163d, MarrNet, pix2surf_2020, Mescheder2019OccNet} and RGBD \cite{yang20173d, kundu20183d, mueller2021completetracking} data does so in isolation, without the human involvement or the interaction.
While challenging, it is arguably more interesting to reconstruct objects in a dynamic setting under severe occlusions from the human.

\subsection{Interaction modelling: Humans and objects with scene context}

\paragraph{Humans in static scenes}
Modelling how humans act in a scene is both important and challenging. Tasks like placement of humans into static scenes~\cite{3d-affordance, PLACE:3DV:2020, Hassan:CVPR:2021}, motion prediction \cite{Cao2020LongtermHM, hassan_samp_2021} or human pose reconstruction~\cite{PROX:2019, chen2019holisticpp, zanfir2018monocular, weng2020holistic, zhang2021learning} under scene constrains,  or learning priors for human-object interactions~\cite{savva2016pigraphs}, have been investigated extensively in recent years.
These methods are relevant but restricted to modelling humans interacting with \emph{static} objects. We address a more challenging problem of jointly tracking human-object interactions in \emph{dynamic} environments where objects are manipulated.

\paragraph{Dynamic human object interactions}
Recently, there has been a strong push on modeling hand-object interactions based on 3D \cite{GRAB:2020, GrapingField:3DV:2020}, 2.5D \cite{Brahmbhatt_2020_ECCV, ContactGrasp2019Brahmbhatt} and 2D \cite{yang2021cpf, ehsani2020force, Corona_2020_CVPR, grady2021contactopt, hasson19_obman} data.
Although powerful, these methods are currently restricted to modelling only \emph{hand-object} interactions. In contrast, we are interested in \emph{full body} capture. Methods for dynamic full body human object interaction approach the problem via 2D action recognition \cite{hu2017jointly_sysu, Liu_2019_NTURGBD120} or reconstruct 3D object trajectories during interactions \cite{dabral2021gravity}. Despite being impressive, such methods either lack full 3D reasoning \cite{hu2017jointly_sysu, Liu_2019_NTURGBD120} or are limited to specific objects \cite{dabral2021gravity}.
\\
More recent work reconstructs and tracks human-object interactions from RGB \cite{sun2021neural} or RGBD streams \cite{su2021robustfusion}, but does not consider contact prediction, thus missing a component necessary for accurate interaction estimates.
\\
Very relevant to our work, PHOSA~\cite{zhang2020phosa} reconstructs humans and objects from a single image. PHOSA uses hand crafted heuristics, instance specific optimization for fitting, and pre-defined contact regions, which limits generalization to diverse human-object interactions. Our method on the other hand learns to predict the necessary information from data, making our models more scale-able.
As shown in the experiments, the accuracy of our method is significantly higher to PHOSA.

\section{\method{} Dataset}
\label{sec:dataset}
\begin{figure*}[t]
    \centering
    \begin{subfigure}{\figTWOsize\linewidth}
      \centering
      \includegraphics[width=\linewidth]{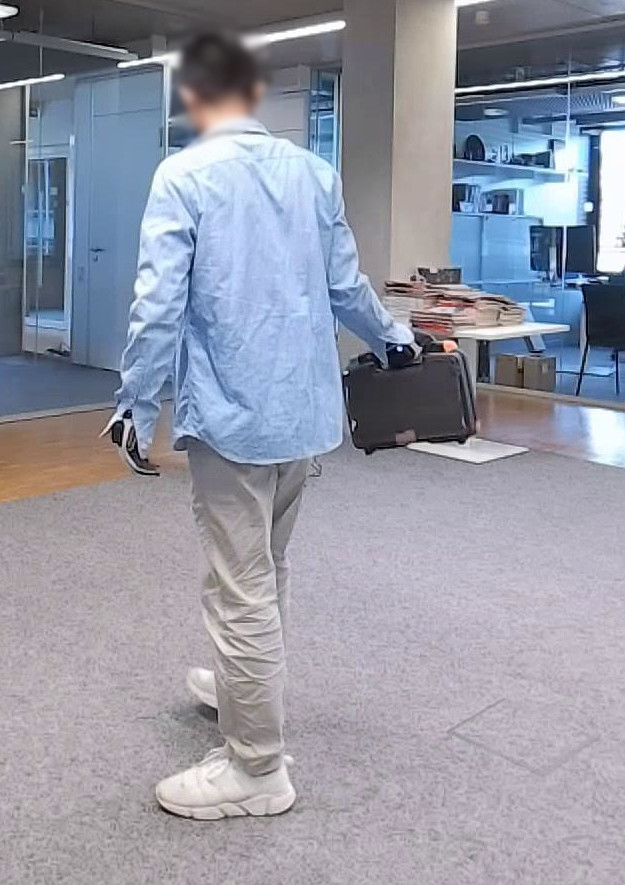}
    \end{subfigure}\hfill%
    \begin{subfigure}{\figTWOsize\linewidth}
      \centering
      \includegraphics[width=\linewidth]{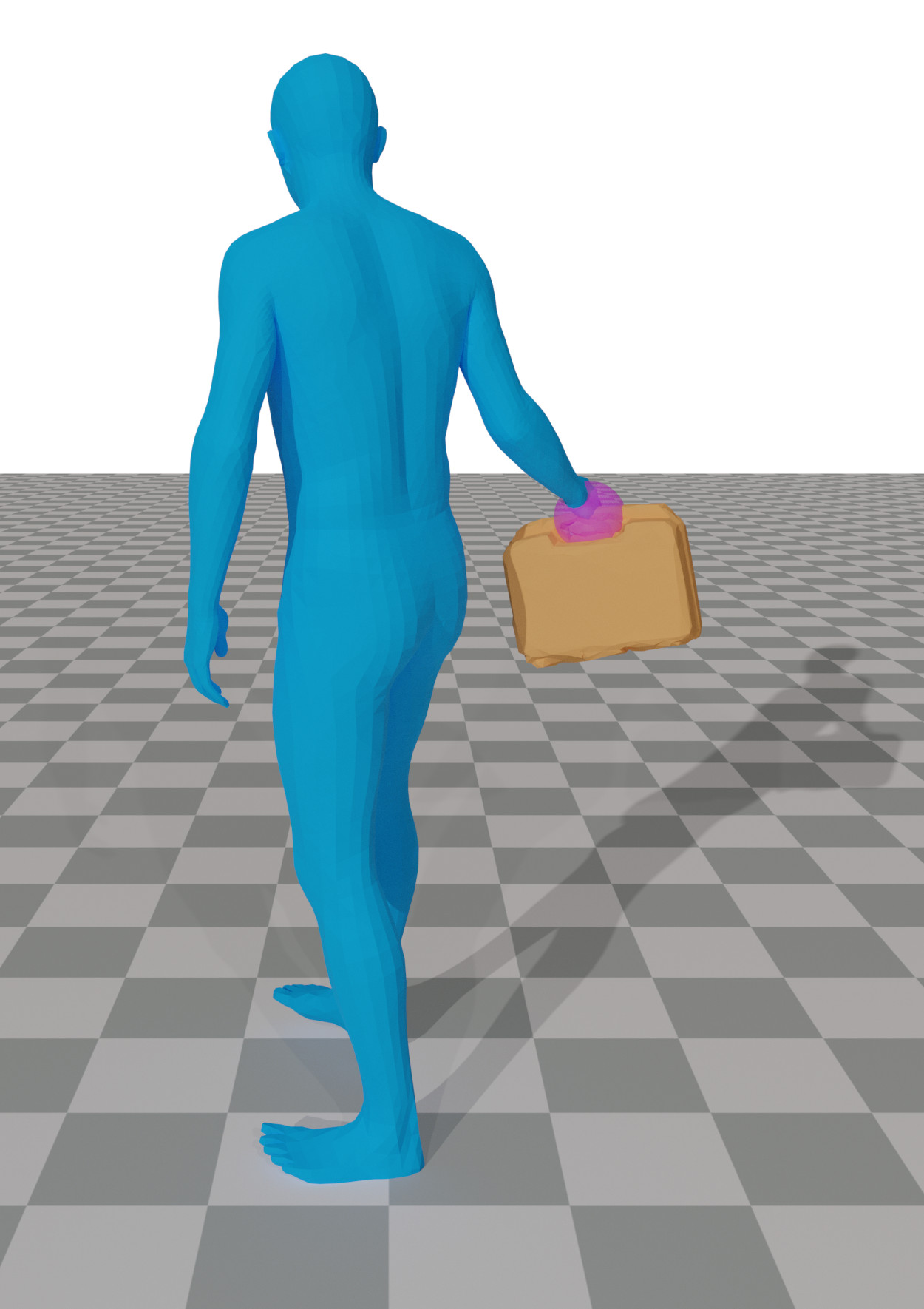}
    \end{subfigure}\hfill%
    \begin{subfigure}{\figTWOsize\linewidth}
      \centering
      \includegraphics[width=\linewidth]{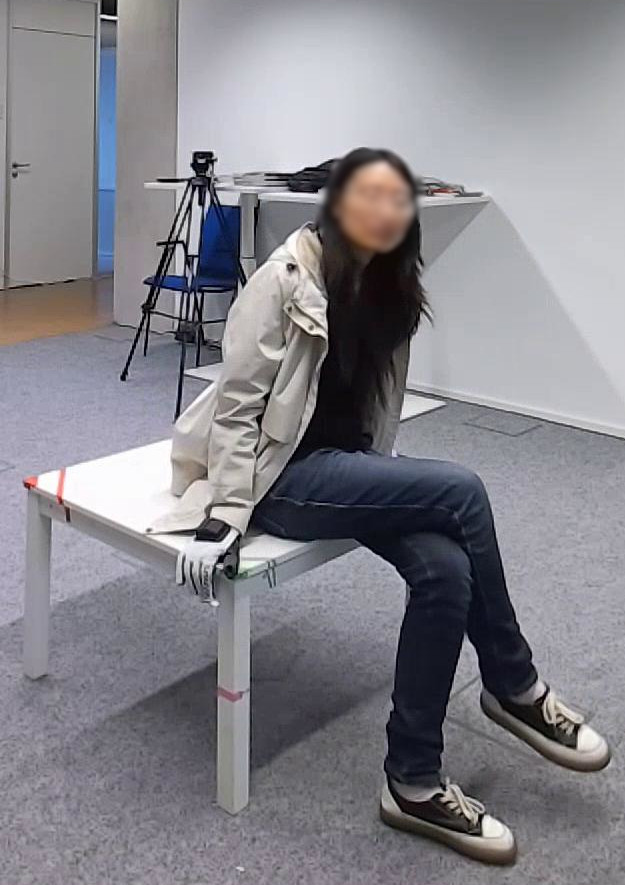}
    \end{subfigure}\hfill%
    \begin{subfigure}{\figTWOsize\linewidth}
      \centering
      \includegraphics[width=\linewidth]{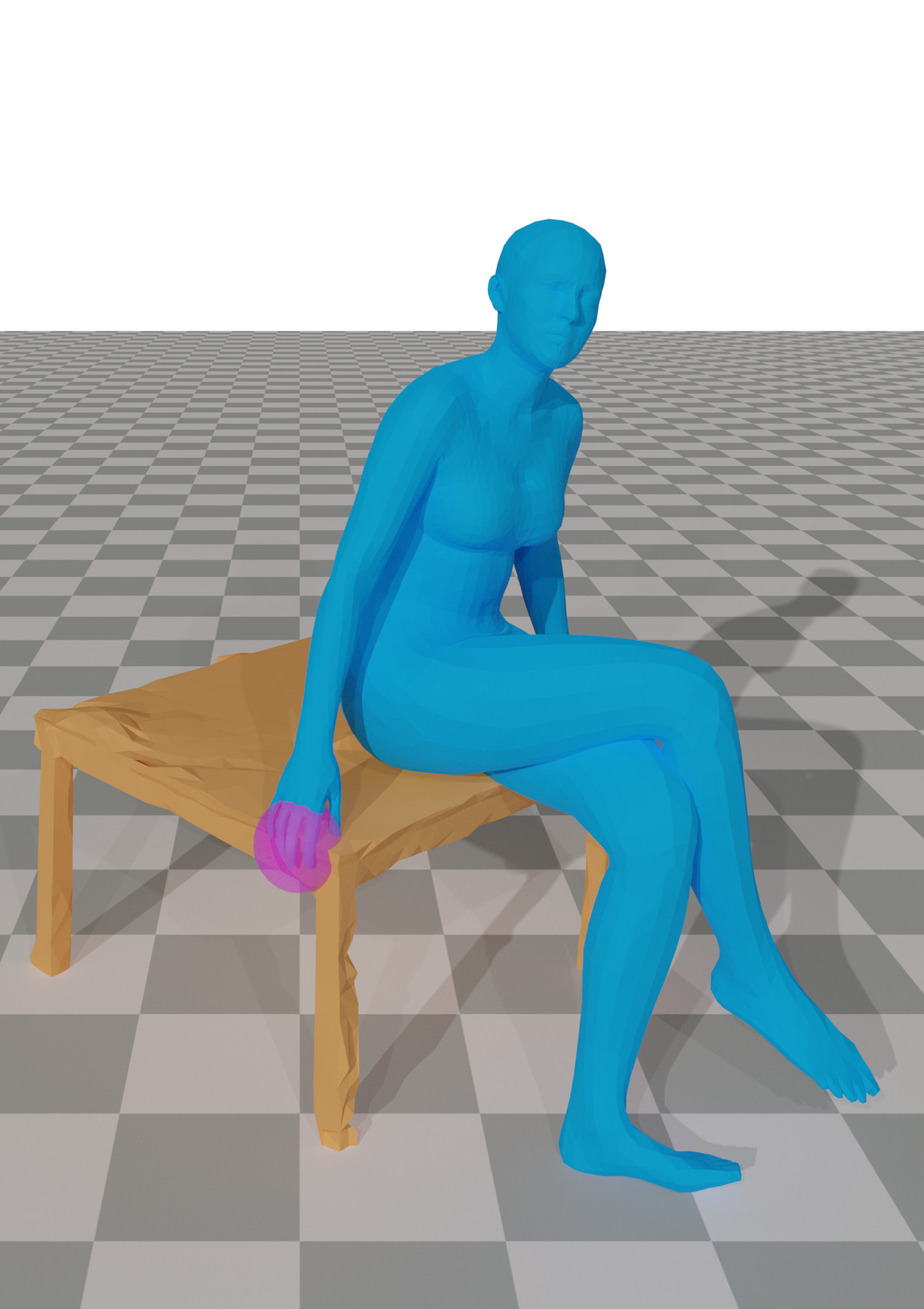}
    \end{subfigure}\hfill%
    \begin{subfigure}{\figTWOsize\linewidth}
      \centering
      \includegraphics[width=\linewidth]{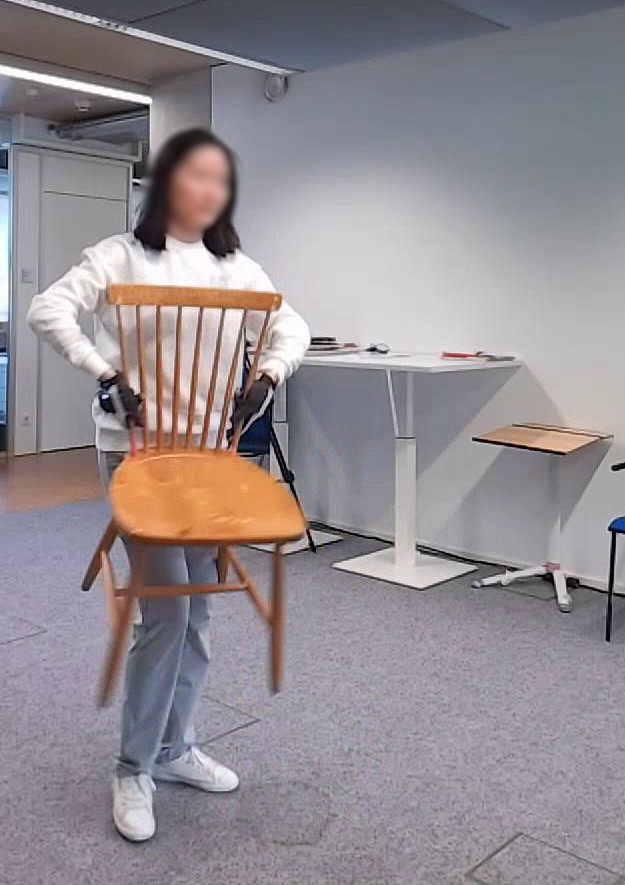}
    \end{subfigure}\hfill%
    \begin{subfigure}{\figTWOsize\linewidth}
      \centering
      \includegraphics[width=\linewidth]{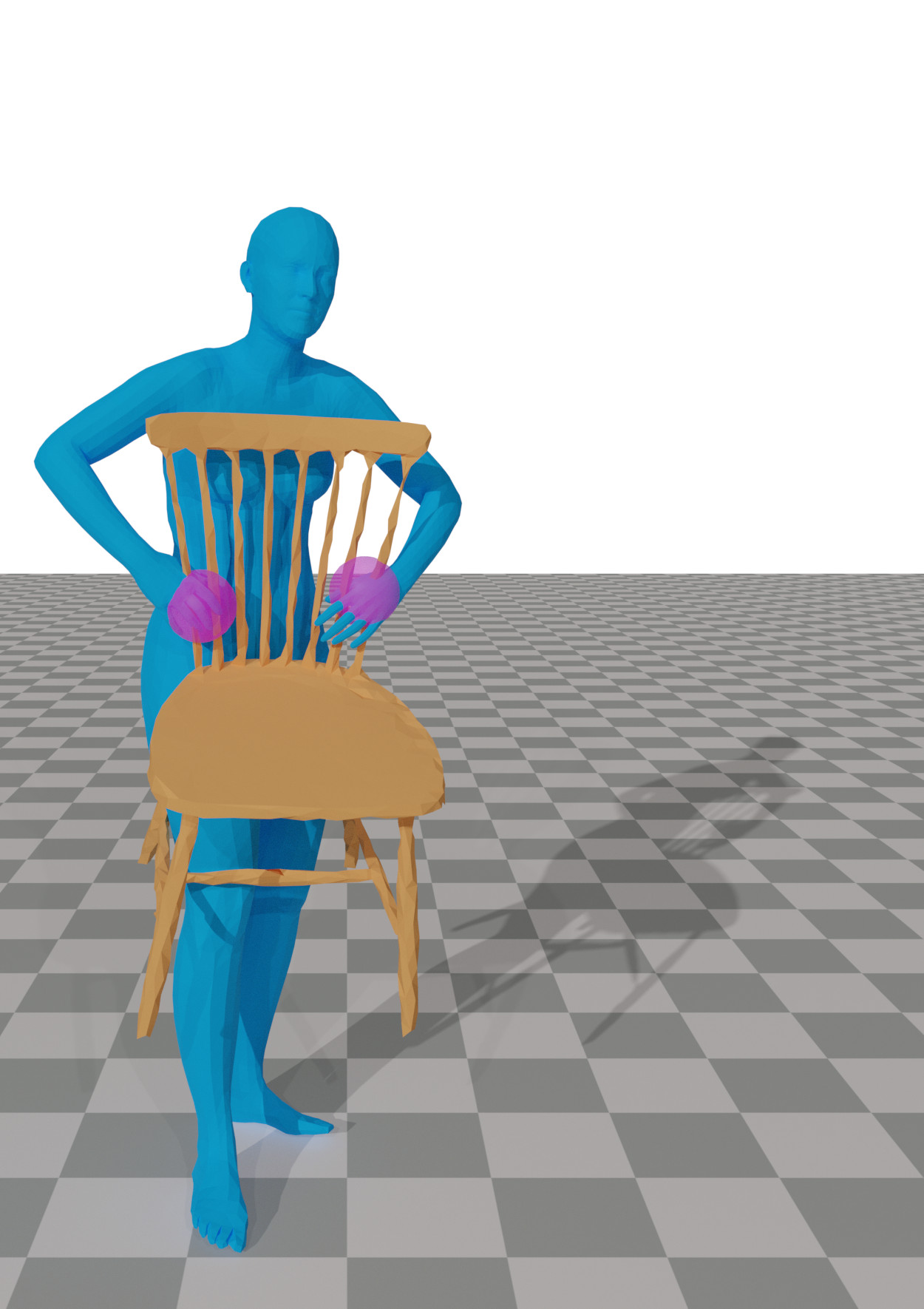}
    \end{subfigure}\hfill%
    \begin{subfigure}{\figTWOsize\linewidth}
      \centering
      \includegraphics[width=\linewidth]{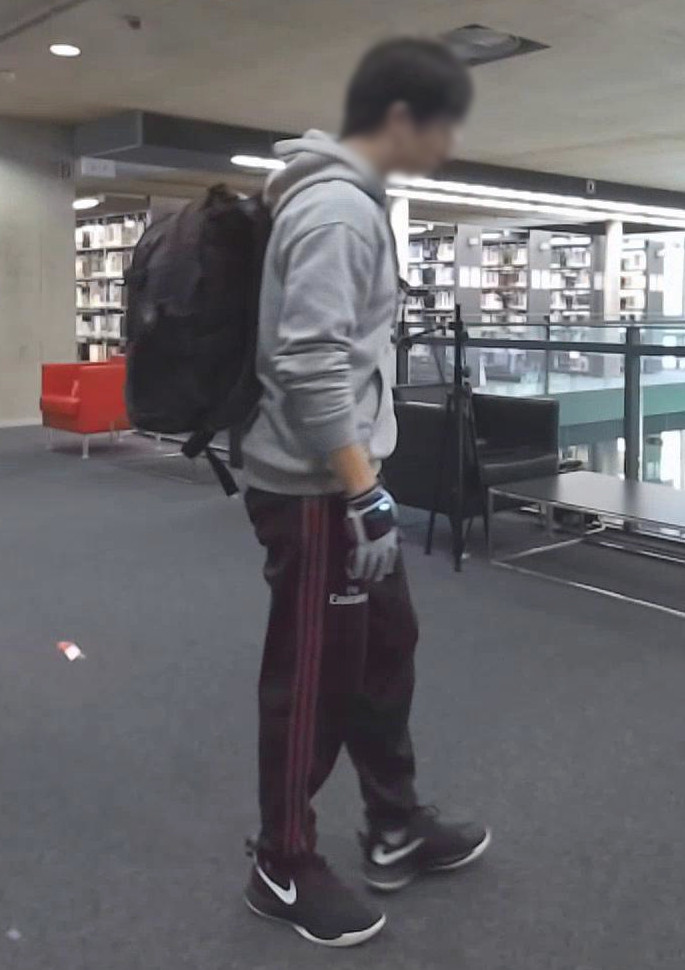}
    \end{subfigure}\hfill%
    \begin{subfigure}{\figTWOsize\linewidth}
      \centering
      \includegraphics[width=\linewidth]{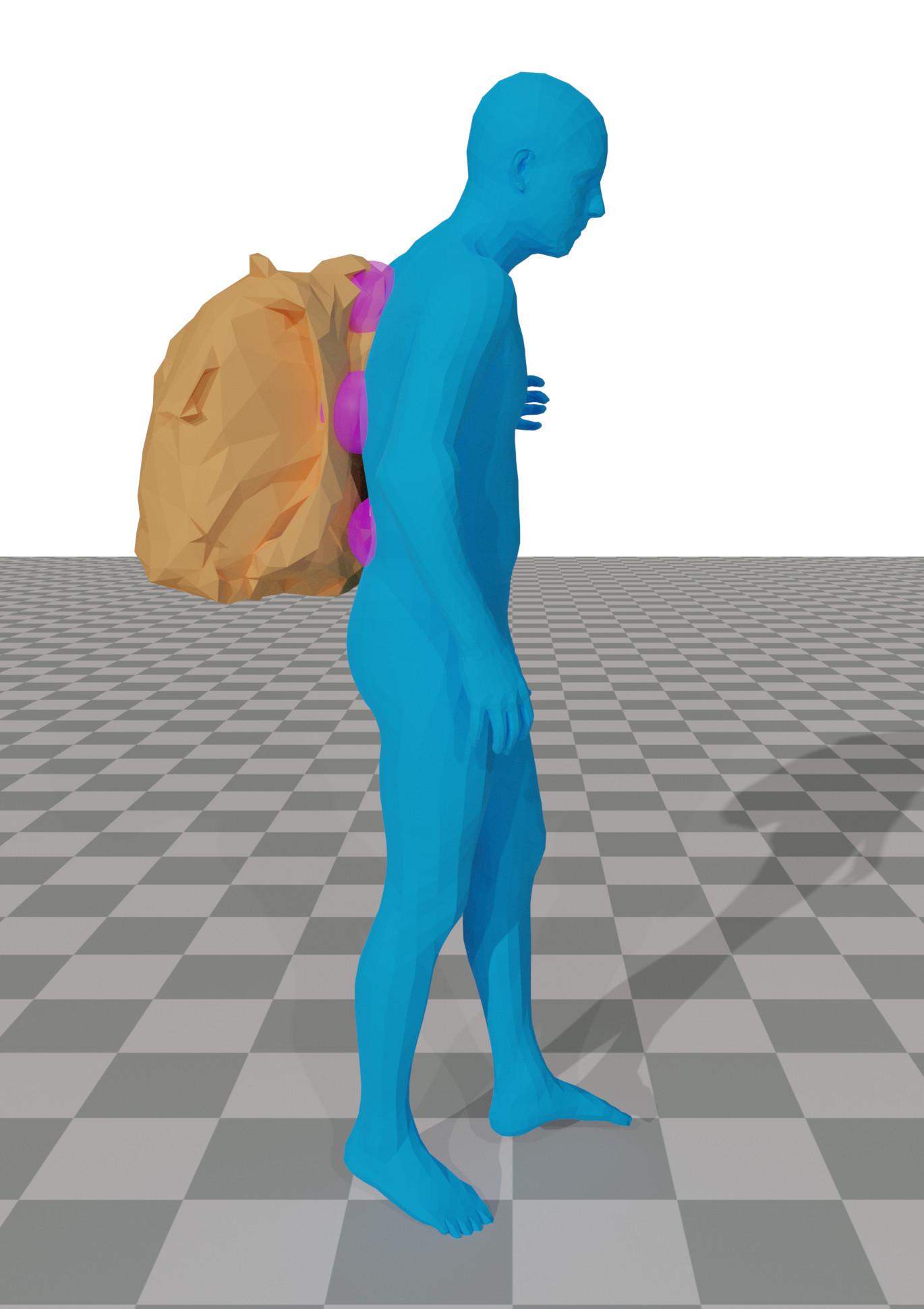}
    \end{subfigure}\hfill%
    \begin{subfigure}{\figTWOsize\linewidth}
      \centering
      \includegraphics[width=\linewidth]{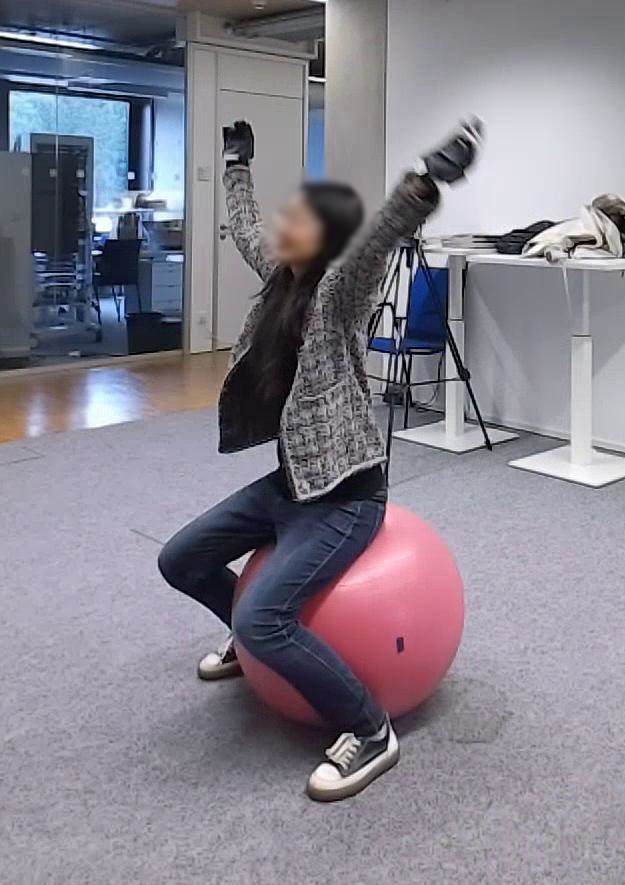}
    \end{subfigure}\hfill%
    \begin{subfigure}{\figTWOsize\linewidth}
      \centering
      \includegraphics[width=\linewidth]{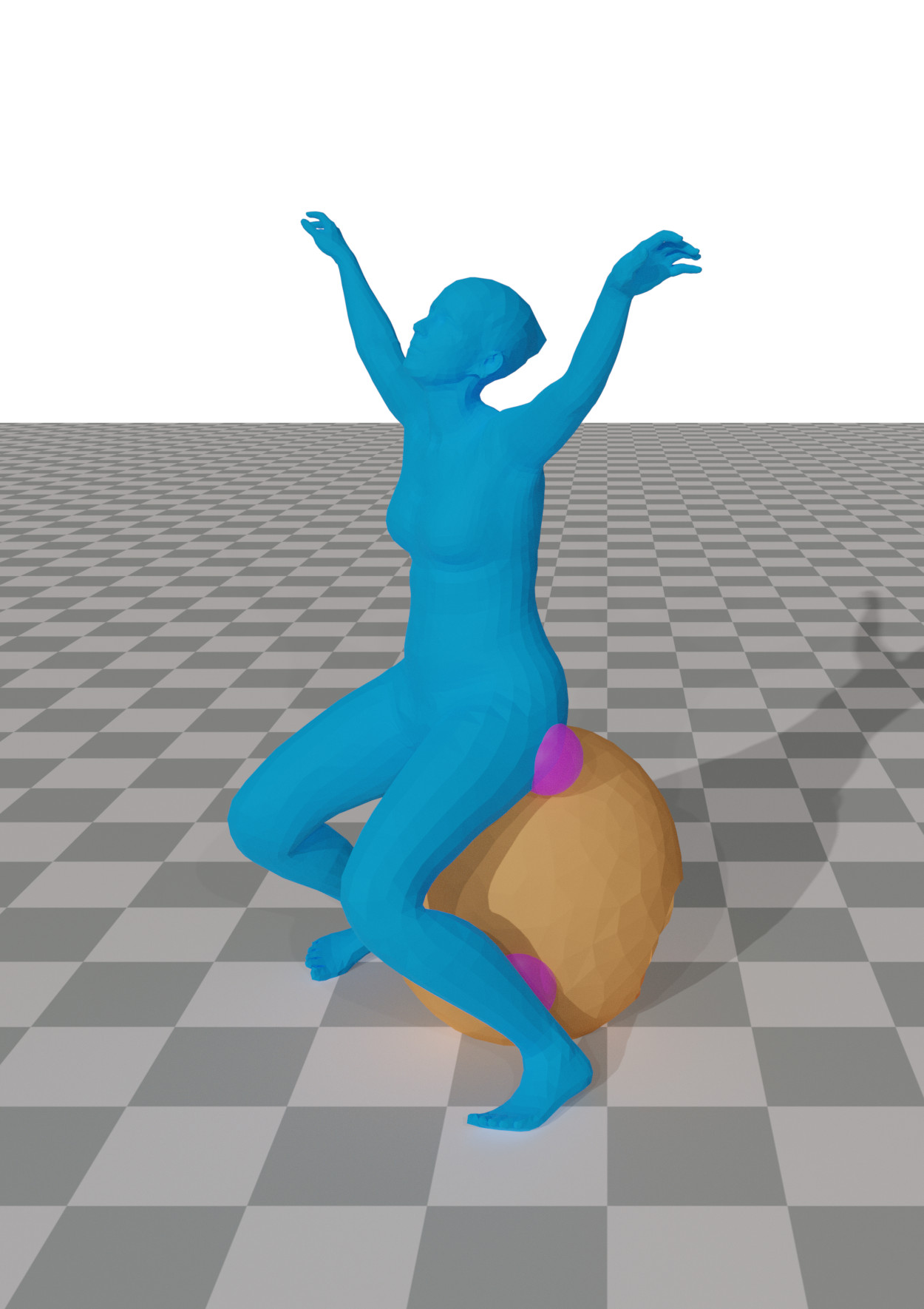}
    \end{subfigure}\hfill%
    \begin{subfigure}{\figTWOsize\linewidth}
      \centering
      \includegraphics[width=\linewidth]{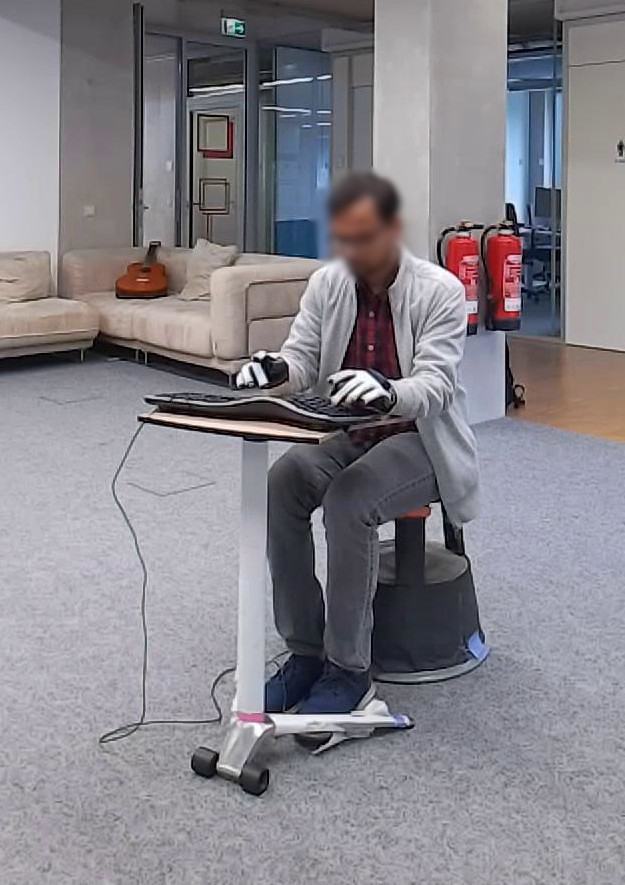}
    \end{subfigure}\hfill%
    \begin{subfigure}{\figTWOsize\linewidth}
      \centering
      \includegraphics[width=\linewidth]{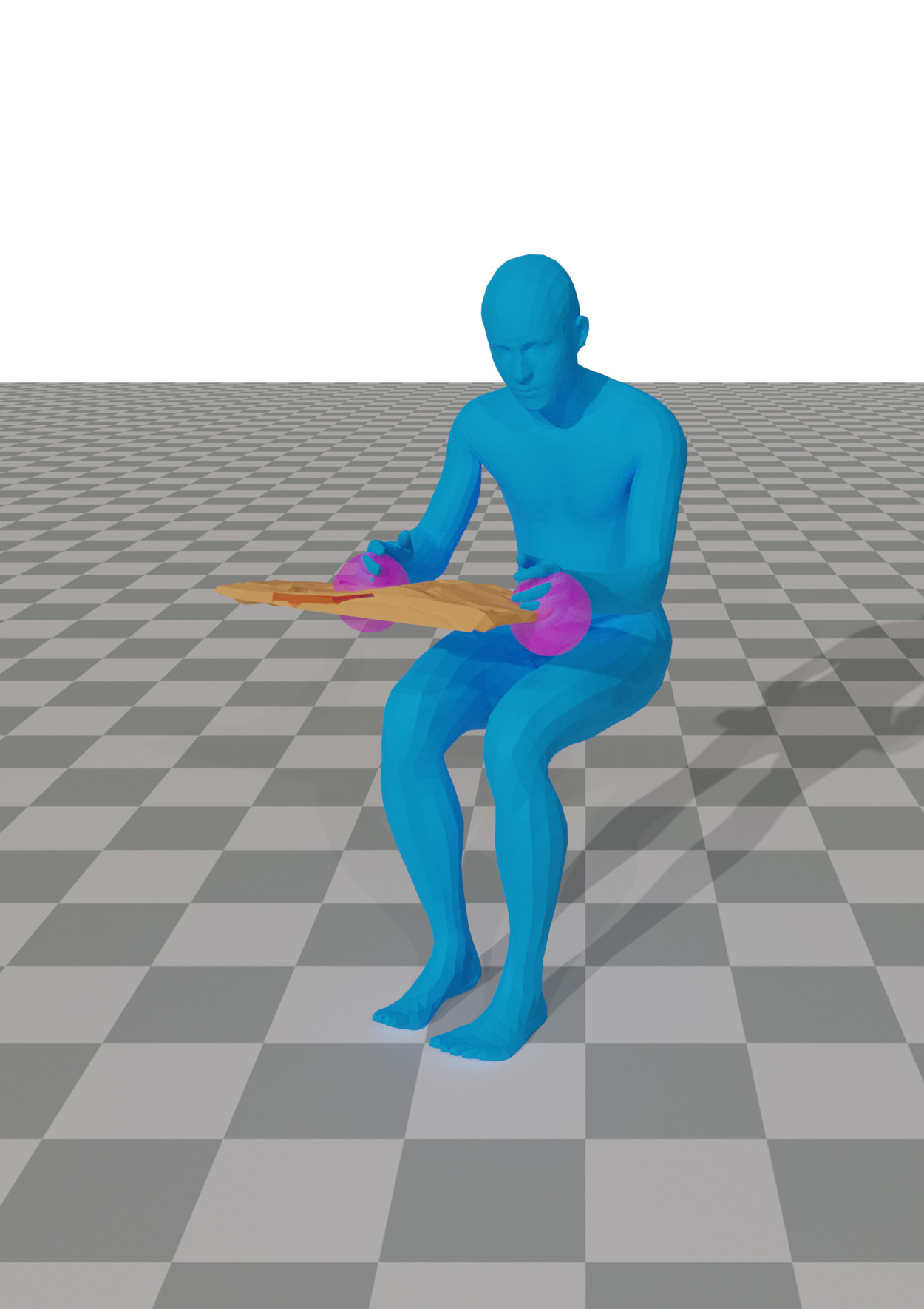}
    \end{subfigure}\hfill%
    \begin{subfigure}{\figTWOsize\linewidth}
      \centering
      \includegraphics[width=\linewidth]{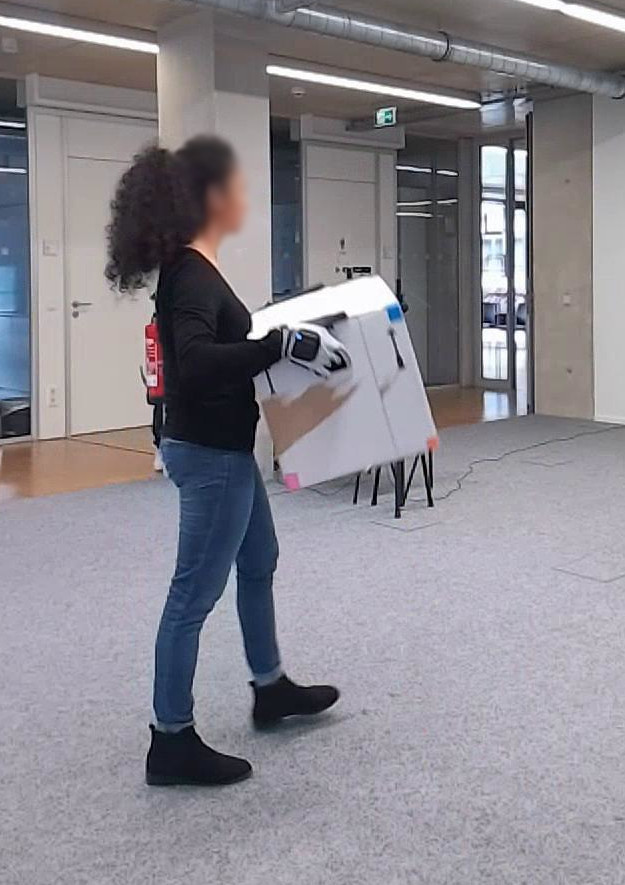}
    \end{subfigure}\hfill%
    \begin{subfigure}{\figTWOsize\linewidth}
      \centering
      \includegraphics[width=\linewidth]{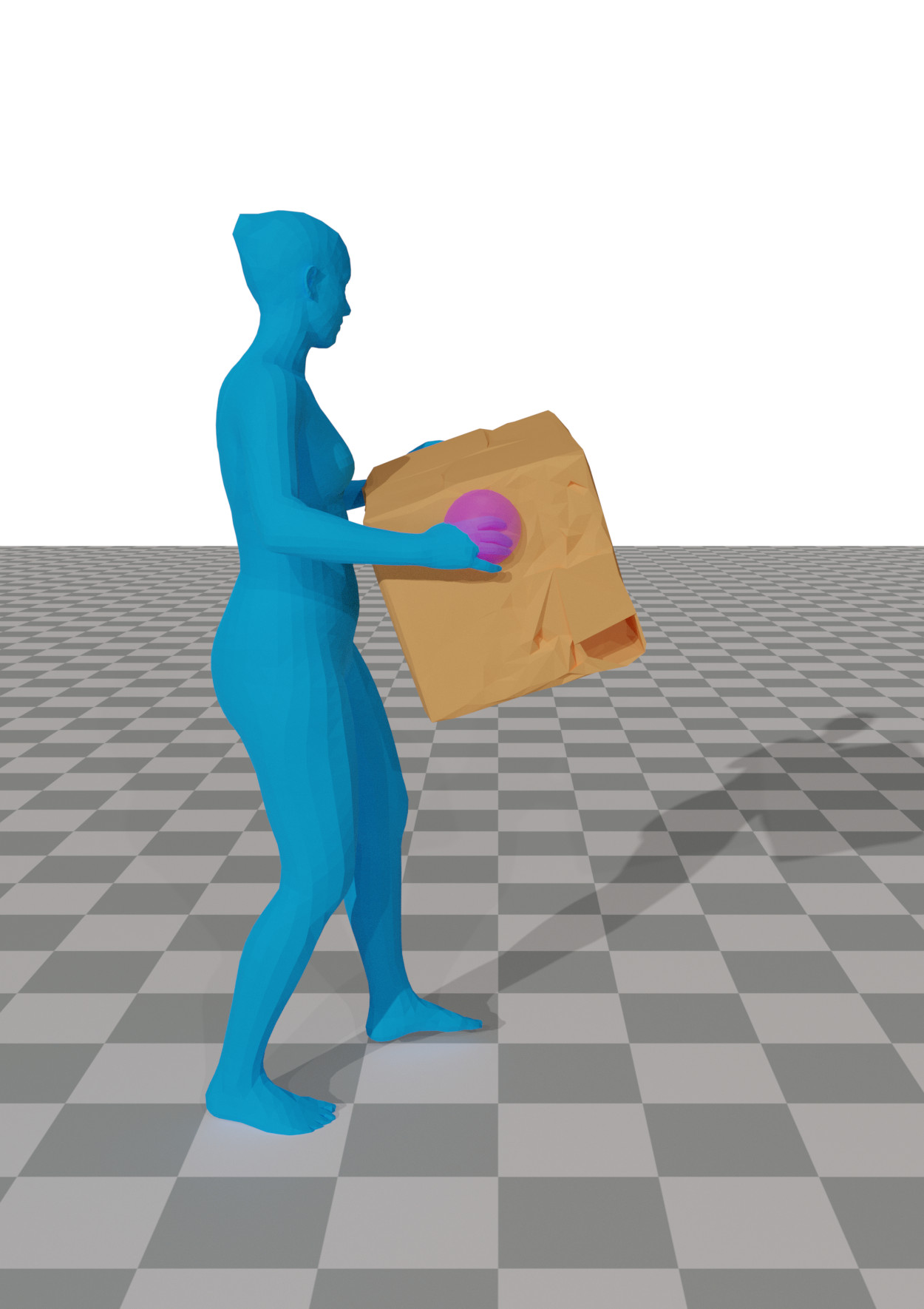}
    \end{subfigure}\hfill%
    \begin{subfigure}{\figTWOsize\linewidth}
      \centering
      \includegraphics[width=\linewidth]{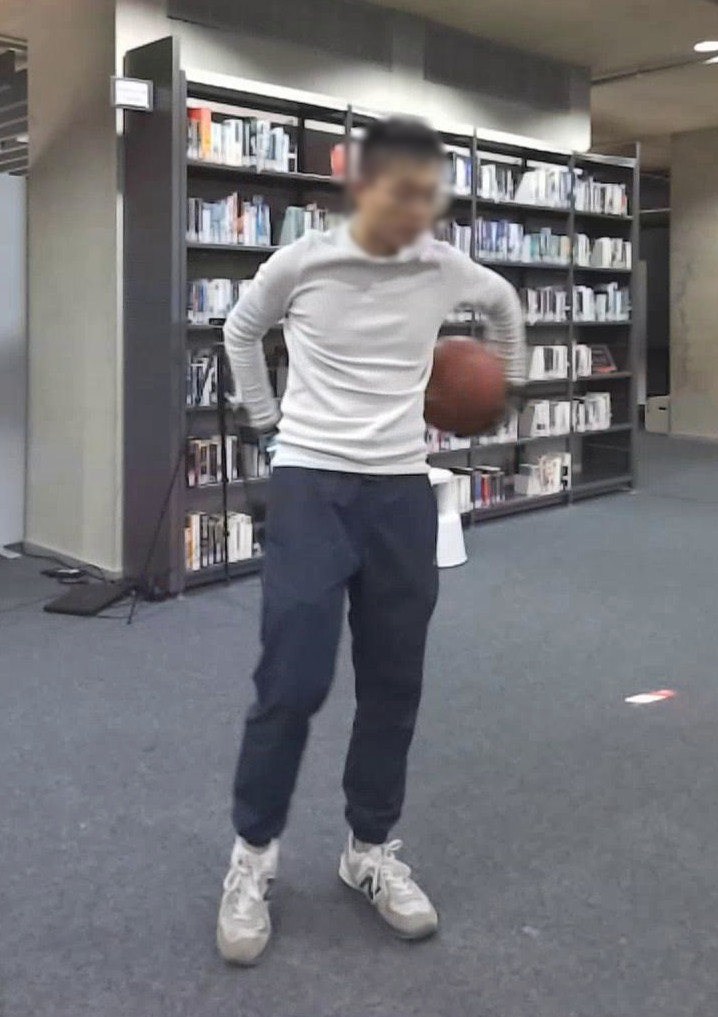}
    \end{subfigure}\hfill%
    \begin{subfigure}{\figTWOsize\linewidth}
      \centering
      \includegraphics[width=\linewidth]{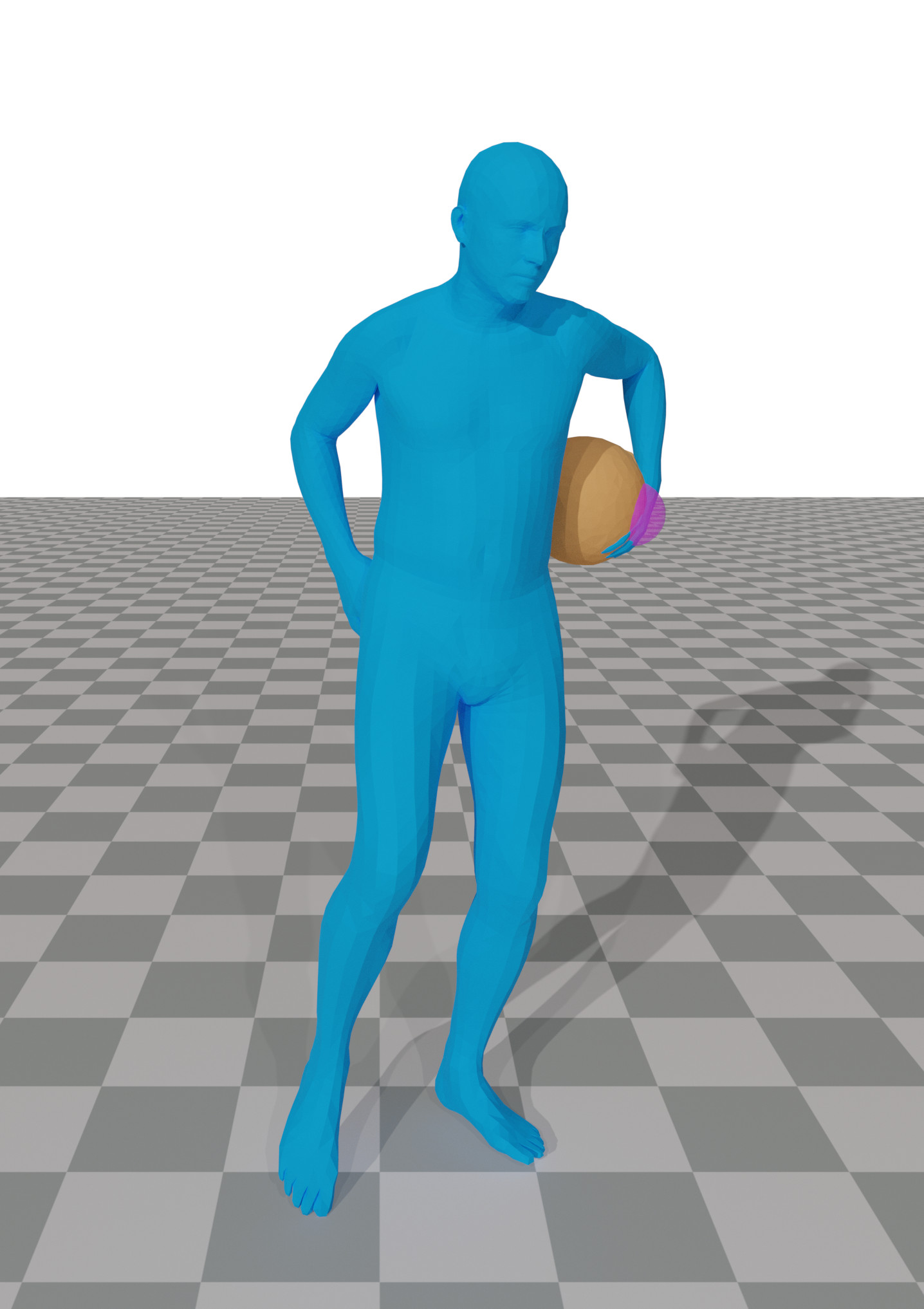}
    \end{subfigure}\hfill%
    \caption{We present \method{} dataset, the \emph{largest} dataset of human-object interactions in natural environments. \method{} contains multi-view RGBD sequences and corresponding 3D object and SMPL fits along with 3D contacts.
  }
  \label{fig:dataset}
\end{figure*}

\begin{table}
  \centering
  \begin{tabular}{@{}lccccc@{}}
    \toprule[1.5pt]
    \specialcell{\bf Dataset} & \specialcell{\bf RGBD} & \specialcell{\bf Hum.} & \specialcell{\bf Ob.Cont.} & \specialcell{\bf Qual.} & \specialcell{\bf Scal.}\\
    \midrule
    NTU~\cite{Liu_2019_NTURGBD120} & \checkmark & Jts. & \text{\sffamily X} & NA & *** \\
    PiGr~\cite{savva2016pigraphs} & \checkmark & Jts. & \text{\sffamily X} & NA & ** \\
    GRAB~\cite{GRAB:2020} & \text{\sffamily X} & \checkmark & \checkmark & *** & *\\
    PROX~\cite{PROX:2019} & \checkmark & \checkmark & Stat. & * & **\\
    \midrule
    Ours & \checkmark & \checkmark & \checkmark & ** & ***\\
    \bottomrule[1.5pt]
  \end{tabular}
  \caption{
    We compare the proposed \method{} dataset with existing ones containing human-object interactions. Our criteria are based on availability of RGB input, 3D human, 3D contact with the object, quality (more stars, better), and scalability to capture at diverse locations (more stars, better). NTU-RGBD~\cite{Liu_2019_NTURGBD120} and PiGraphs~\cite{savva2016pigraphs} do not provide full 3D human and object contacts and are hence unsuitable for modelling dynamic 3D interactions. GRAB~\cite{GRAB:2020} uses a marker based capture system and hence contains the highest quality data but this also makes it difficult to scale. PROX~\cite{PROX:2019} is easier to scale as it uses a single Kinect based capture setup (although, scene needs to be pre-scanned) but this reduces the overall quality. More importantly it does not contain dynamic interactions.
    Ours is the first dataset that captures dynamic human-object interactions in diverse environments.}
  \label{tab:comparison_dataset}
\end{table}

We present \method{} dataset, the \emph{largest} dataset of human-object interactions in natural environments, with 3D human, object and contact annotation, to date. See~\cref{tab:comparison_dataset} for comparison with other datasets.
Our dataset contains multi-view RGBD frames, with accurate pseudo-ground truth SMPL \cite{smpl2015loper}, object fits, human and object segmentation masks, and contact annotations. 
\paragraph{Recording multi-view RGBD data}
We setup and calibrate 4 Kinects at 4 corners of our square recording volume where all interactions are performed by 8 subjects (5 male, 3 female). Interactions are captured at 5 disparate indoor locations with 20 commonly used, yet diverse objects: 5 different boxes, 2 chairs, 2 tables, crate, backpack, trashcan, monitor, keyboard, suitcase, basketball, exercise ball, yoga mat, stool and a toolbox. We include common interactions such as lifting, carrying, sitting, pushing and pulling with hands and feet, as well as free interactions. See our supplementary video for sample sequences.
In total, our dataset contains 10.7k frames for training and 4.5k frames for testing respectively.

\paragraph{Human segmentation and SMPL fitting}
We segment the human in our images by running DetectronV2~\cite{wu2019detectron2} followed by manual correction with \cite{fbrs2020} on the segmentation masks. These masks are then used to segment multi-view depth maps and lift human point clouds from 2D to 3D. We use FrankMocap~\cite{rong2021frankmocap} to initialize SMPL's pose from the images and then use instance specific optimization~\cite{alldieck2019learning} to fit the SMPL model to the segmented human point cloud. For more accurate fitting, we additionally obtain the SMPL shape parameters of each subject from 3D scans using \cite{bhatnagar2020ipnet}. We report a chamfer error of 1.80cm between the segmented kinect point cloud and our SMPL fits.

\paragraph{Object segmentation and fitting}
To obtain object segmentation, we pre-scan objects using a 3D scanner~\cite{treedys, agisoft}. We then use multi-view object keypoints, marked manually by AMT~\cite{amt} annotators in images, to optimize the 6D pose of the pre-scanned object mesh to the given frame. We obtain the chamfer error of 2.42cm between the segmented Kinect point cloud and object fit.
The segmentation masks are then obtained by projecting fitted object meshes to the images.

\paragraph{Contact annotation} Based on the pseudo-GT SMPL and object fits as described above, we automatically detect contacts if a point on the human surface (registered SMPL) is closer than 2cm to the object surface. For every object point, we store a binary contact label (whether there is a  contact or not) and correspondence to the human (contact location on the surface).

\noindent See supplementary for more details on data acquisition.

\paragraph{How will this dataset be useful to the community?} We devote significant effort in recording the \emph{largest}, so far, dataset of natural, full body, day-to-day human interactions with common objects in different natural environments.
We propose following challenges with \method{} dataset:
\begin{itemize}
    \item {\bf Tracking human-object interactions.} Track humans and objects using multi-view RGBD data. This can further be extended to track with just multi-view RGB, no-depth, and eventually just a single camera.
    \item {\bf Reconstruction from a single image.} Joint 3D reconstruction of 3D humans and objects from a single RGB image. Currently, there is no dataset that can be used for benchmarking let alone to learn such a model.
    \item {\bf Pose and shape estimation.} Benchmarking pose and shape estimation methods in challenging natural environments where the person is heavily occluded by the interacting object.
\end{itemize}
Apart from these tasks, the research community is free to explore other applications of the \method{} dataset.

\section{Method: Tracking human, object and contacts}
\label{sec:method}
\begin{figure*}[t]
  \centering
  \fbox{\includegraphics[width=\figTHREEsize\linewidth]{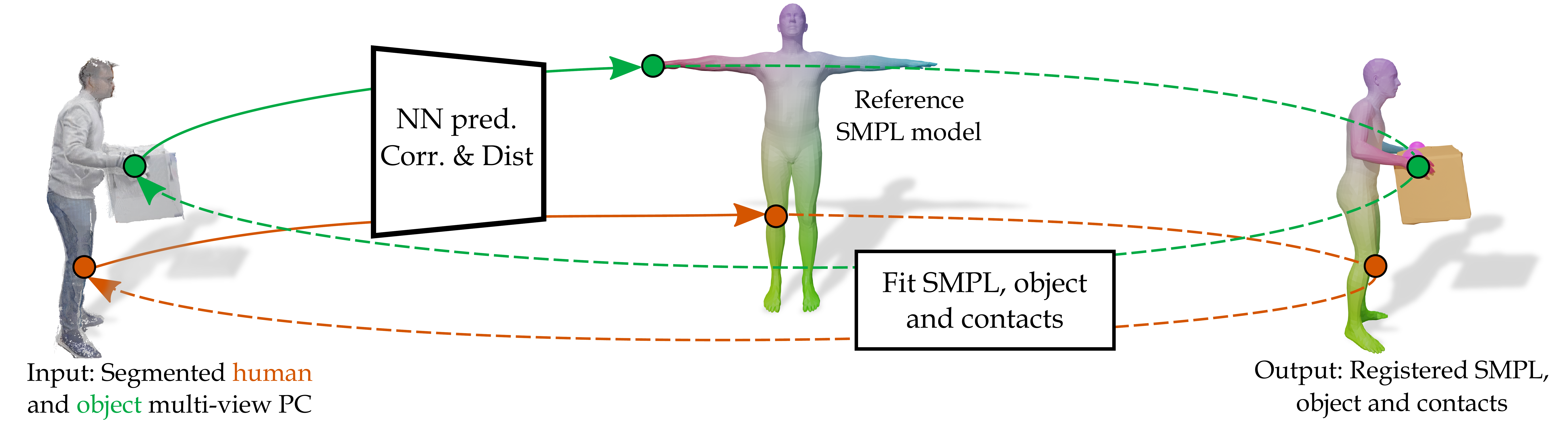}}
  \caption{Given a sequence of multi-view images, we track the human and the object using SMPL and a template object mesh. We lift the segmented multi-view RGBD frames to 3D and obtain a human and object point cloud. 
  As shown here, our network predicts correspondences between the human point cloud and the body model, which allows us to fit SMPL. We also predict correspondences from the object to  the body model, thus allowing us to model contacts.
  Our network predictions (see~\cref{sec:method}) allow us to register SMPL and the object mesh to a video, making an accurate joint tracking of the human and object possible.
  }
  \label{fig:overview}
\end{figure*}

We present \method{}, a method to jointly track humans, objects and their interactions (represented as surface contacts) based on multi-view RGBD input.
We formulate our method as an extended per-frame registration problem: we register the human (using SMPL~\cite{smpl2015loper}) and the object (using its pre-scanned object mesh), and predict contacts as correspondences between SMPL and object meshes. See~\cref{fig:overview} for the overview of our method.
\\
Our formulation must obey three properties, (i) the SMPL model $\smpl(\cdot)$ should fit the human in the multi-view input, (ii) the object mesh $\mat{W}^o$ should fit the input object and, (iii) SMPL model and object should satisfy contacts.
\\
To facilitate joint reasoning of the human, object and contacts directly in 3D, we lift the human $\set{S}^h$, and object $\set{S}^o$ point clouds to 3D using multi-view depth and semantic segmentation.
Our joint formulation fits SMPL $\smpl(\cdot)$ and the object $\mat{W}^o$ to multi-view RGB-D data at each time step, using explicit contacts. This takes the following form
\begin{equation}
\label{eq:overall_objective}
\begin{split}
    E(\pose, \shape, \mat{R}^o, \vect{t}^o) = & E^\text{SMPL}(\set{S}^h, \smpl(\pose, \shape)) + E^\text{obj}(\set{S}^o, \mat{W}^o) + \\ & E^\text{contact}(\mathbbm{1}^c\mat{W}^o, \smpl(\pose, \shape)).
\end{split}
\end{equation}
The SMPL model is parameterized by pose $\pose$, and shape $\shape$. For notation brevity, we include the global SMPL translation into the pose parameters. We assume the template object $\mat{W}$ be rigid and only estimate the rotation $\mat{R}^o$, and translation $\vect{t}^o$, to fit the object mesh $\mat{W}^o = \mat{R}^o \mat{W} + \vect{t}^o$, to the object point cloud.
\\
The indicator matrix, $\mathbbm{1}^c$, selects the vertices on the object mesh $\mat{W}^o$, that are in contact with the SMPL model. 
This ensures that contact locations on the object and the human mesh adequately align in 3D. 
\\
The term $E^\text{SMPL}(\set{S}^h, \smpl(\pose, \shape))$ is designed to accurately fit SMPL to the human point cloud $\set{S}^h$. The term $E^\text{obj}(\set{S}^o, \mat{W}^o)$ is designed to fit the object mesh to the object point cloud and $E^\text{contact}(\mathbbm{1}^c\mat{W}^o, \smpl(\pose, \shape))$ ensures that contacts between the human and object match (align). We explain each term in detail next.

\subsection{Fitting human model to the human point cloud}
\label{sec:SMPL_fitting}
Fitting SMPL to the human point cloud $\humanpc$ requires, (i) that distance between the SMPL model and the human point cloud should be minimized and (ii) the correct SMPL parts fit the corresponding body parts of the point cloud. The latter is important to avoid degenerate cases such as $180^\circ$ flipped fitting, where the left hand is erroneously matched to the right side of the body or vice-versa~\cite{bhatnagar2020ipnet}.
With these considerations, we design our SMPL fitting objective as:
\begin{equation}
\label{eq:SMPL_fitting}
    E^\text{SMPL} = \funct{d}(\set{S}^h, \smpl(\pose, \shape)) + E^\text{corr} + E^\text{reg},
\end{equation}
where $\funct{d}(\set{S}^h, \smpl(\pose, \shape))$ minimizes the point-to-mesh distance between the input human point cloud $\set{S}^h$ and the SMPL model.
To avoid sub-optimal local minima during fitting~\cite{bhatnagar2020ipnet, bhatnagar2020loopreg}, we train a neural network that predicts dense correspondences from the input to the SMPL model. This ensures that correct SMPL parts explain corresponding input regions, using the term $E^\text{corr}$. 
\\
Specifically, we train an encoder network similar to~\cite{ih_chibane2020,chibane2020ndf} that takes the segmented and voxelized human $\humanpc$ and object $\objectpc$ point cloud as inputs, and generates a voxel aligned grid of features $\mat{F} = \net^\text{enc}(\humanpc, \objectpc)$.
We then sample $N$ 3D query points, $\{\vect{p}_1, \dots, \vect{p}_N\}, \vect{p}_i \in \mathbb{R}^3$ and for each point $\vect{p}_i=(x,y,z)$ obtain the corresponding point feature $\mat{F}_i=\mat{F}(x,y,z)$.
We pass this point feature through a decoder network $\net^\text{udf}$, to predict the unsigned distance to object and human surfaces, $u_i^o, u_i^h = \net^\text{udf}(\mat{F}_i), u_i^o, u_i^h \in \mathbb{R}$, respectively. We use a second decoder network $\net^\text{corr}$, to predict the correspondence of point $\vect{p}_i$ to the SMPL model, $\vect{c}_i = \net^\text{corr}(\mat{F}_i), \vect{c}_i \in \mathbb{R}^3$.
\\
$E^\text{corr}$ enforces that the distance between the input point $\vect{p}_i$ and the corresponding point $\vect{c}_i$ after transforming it with the SMPL model is same as the distance predicted by the network $u^h_i$. Under a slight abuse of notation we use $\smpl(\vect{c}_i, \pose, \shape)$ to transform $\vect{c}_i$ with the SMPL function.
\begin{equation}
\label{eq:SMPL_fitting_correspondences}
    E^\text{corr} = \sum_{i=1}^{N} ||\vect{p}_i - \smpl(\vect{c}_i, \pose, \shape)|_2 - u^h_i|.
\end{equation}
If the correspondences predicted by the network $\vect{c}_i$ deviate from the SMPL surface, these cannot be skinned using the SMPL model as its function is only defined on the body surface. To alleviate this issue, we use the LoopReg~\cite{bhatnagar2020loopreg} formulation that allows us to pose and shape off-the-surface correspondences as well.
\\
The final term $E^\text{reg} = E^\text{J2D} + E^{\pose} + E^{\shape}$, adds regularisation for SMPL joints, $E^\text{J2D} = \sum_{k=1}^K |\pi_k\smpl^J(\pose, \shape) - \mat{J}_{2D}^k |_2$, where $\pi_k$ is the camera projection matrix of camera $k$, $\smpl^J(\cdot)$ are the 3D body joints and $\mat{J}_{2D}^k$ are the 2D joints detected in the $k^{th}$ Kinect image. $E^{\pose}$ and $E^{\shape}$ are regularisation terms on SMPL pose and shape similar to~\cite{bogo2016smplify}.

\subsection{Fitting the object mesh to the object point cloud}
\label{sec:object_fitting}
In order to fit the object mesh, we must ensure that distance from the input object point cloud to the object mesh is minimized. Minimizing this one-sided distance is necessary but not sufficient.
Since severe occlusions are common in our interaction setting, large parts of object might be missing from the object point cloud, making fitting difficult.
To alleviate this issue we must also ensure that all the vertices of the object mesh are correctly placed w.r.t. the input, even when the point cloud is incomplete.
To do so, we take the object mesh vertices $\vect{v}^o_j \in \mat{W}^o, j \in \{1, \dots, L\}$ and obtain the corresponding point feature $\mat{F}_j$, same as~\cref{sec:SMPL_fitting}, where $L$ is the number of object mesh vertices.
We then obtain the unsigned distances to the object and human surfaces using the point feature $u_j^o, u_j^h = \net^\text{udf}(\mat{F}_j)$. Since $\vect{v}^o_j$ is a vertex on the object mesh, its distance to the object surface $u^o_j$ must be zero for a correct fit. This allows us to accurately fit the object vertices to the point data even when corresponding parts are missing from the object point cloud.
\begin{equation}
\label{eq:object_fitting}
    E^\text{obj} = \funct{d}(\objectpc, \mat{W}^o) + \sum_{j=1}^L |u^o_j|,
\end{equation}
where, $\funct{d}(\objectpc, \mat{W}^o)$ minimizes the point-to-mesh distance between the object point cloud and the object mesh, and the term $\sum_{j=1}^L |u^o_j|$ uses implicit unsigned distance prediction to reason about missing object parts.

\begin{figure}[t!]
    \centering
    \begin{subfigure}{0.4\linewidth}
      \centering
      \caption{Input PC}
      \includegraphics[width=\linewidth]{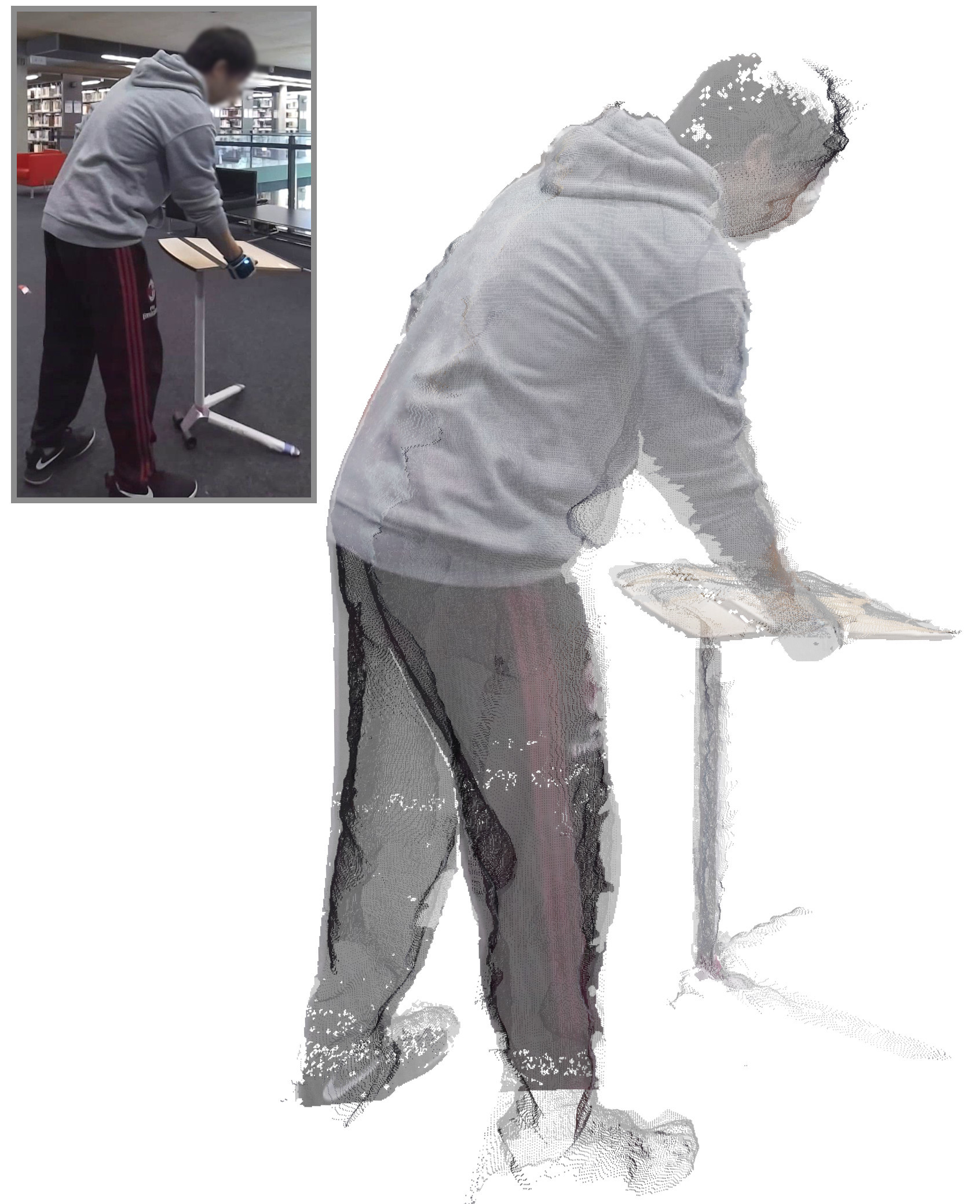}
    \end{subfigure}\hfill%
    \begin{subfigure}{\figFOURsize\linewidth}
      \centering
      \caption{Our w/o ori.}
      \includegraphics[width=\linewidth]{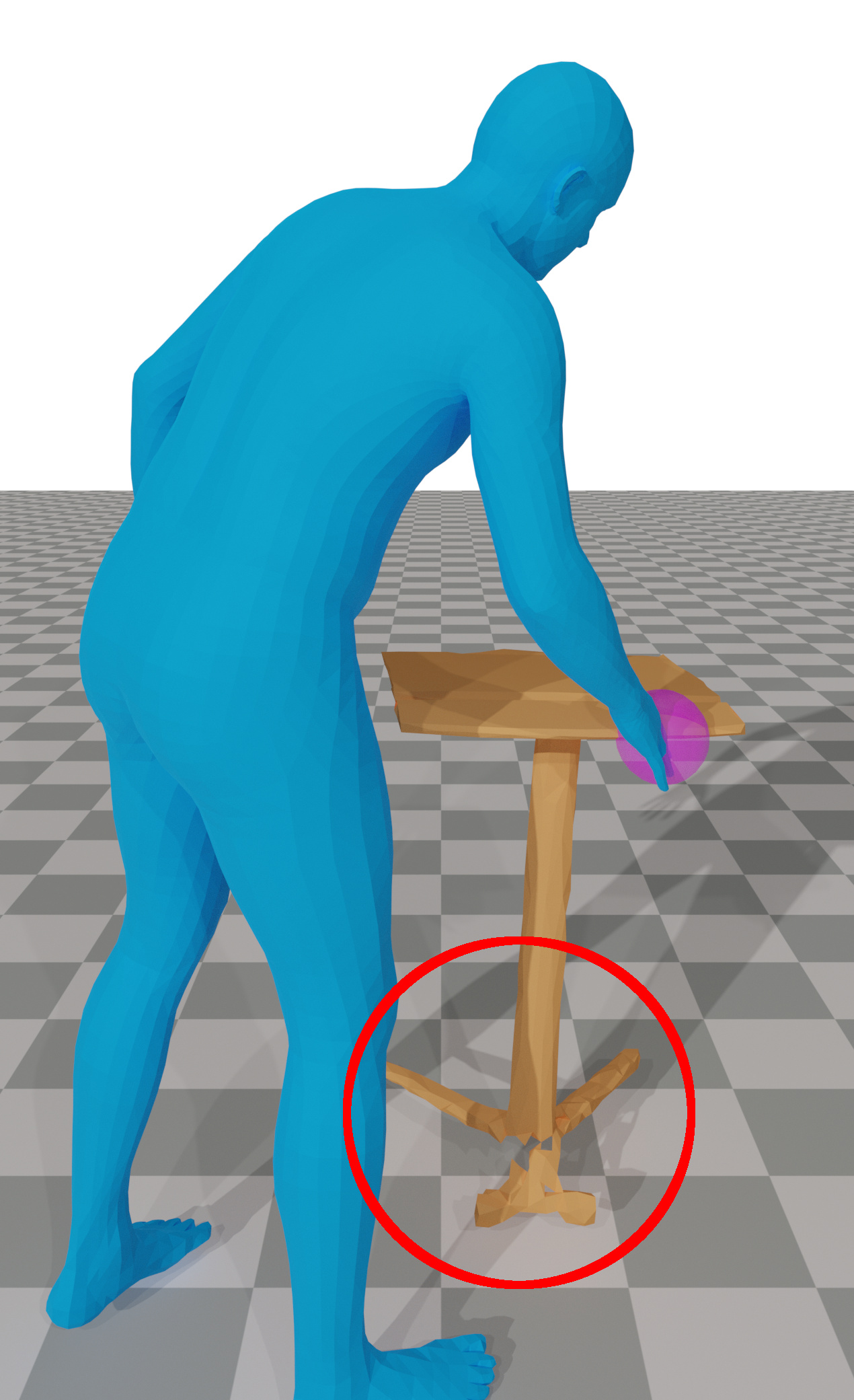}
    \end{subfigure}\hfill%
    \begin{subfigure}{\figFOURsize\linewidth}
      \centering
      \caption{Ours}
      \includegraphics[width=\linewidth]{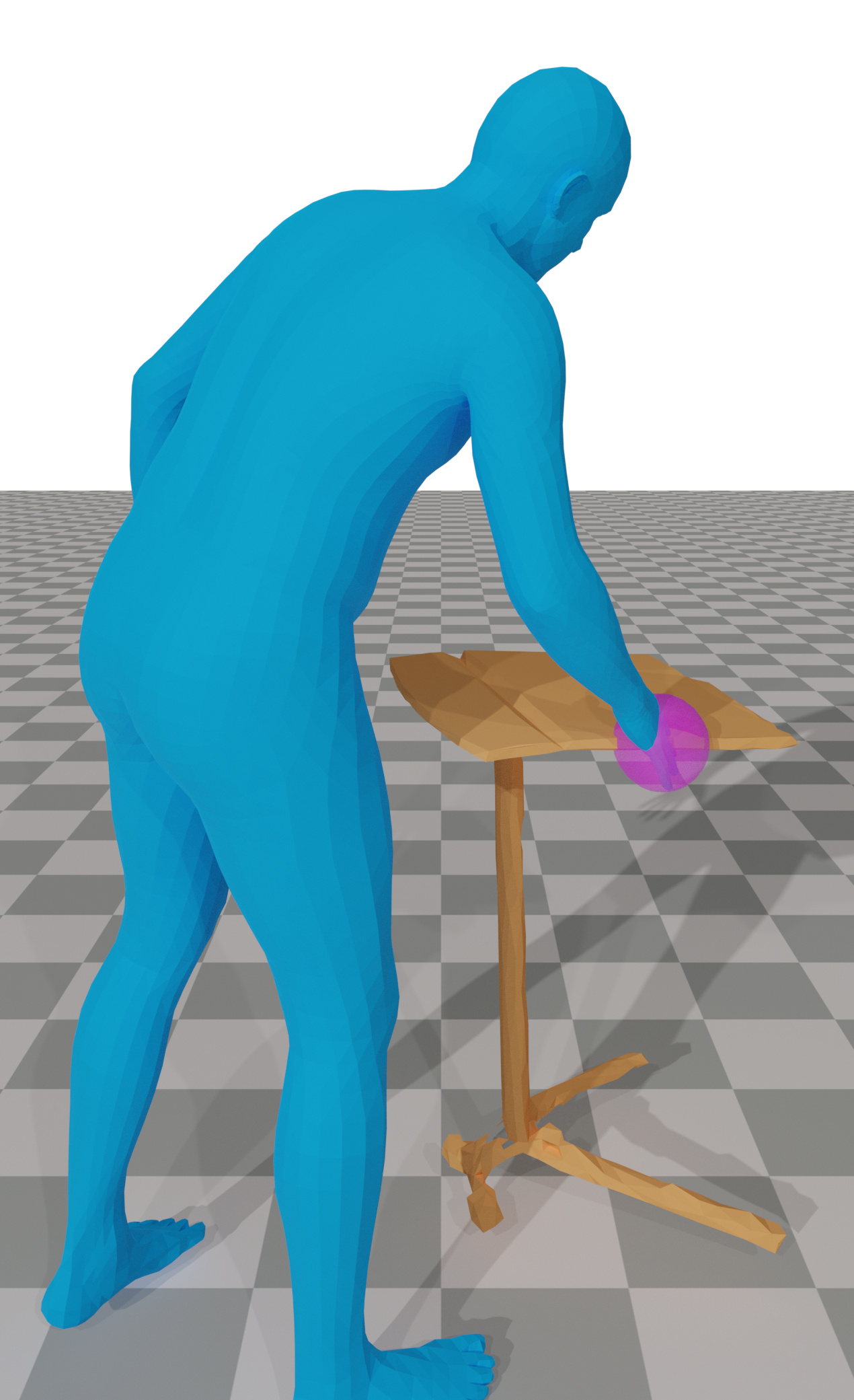}  
    \end{subfigure}
    \caption{We show that our network predicted orientation is important for accurate object fitting. Without our orientation prediction the fitting gets stuck in a local minima.}
\label{fig:importance_of_initialization}
\end{figure}
\paragraph{Predicting object orientation.} Although the terms in~\cref{eq:object_fitting} minimize the bi-directional distance between the object point cloud and the object mesh, they do not guarantee that parts of the object point cloud are explained by the semantically corresponding parts on the object mesh, e.g. in~\cref{fig:importance_of_initialization}, the legs of the table are not aligned correctly.
This issue can be fixed if we obtain the global object orientation during fitting. We represent the orientation of the object with the principal components obtained by running PCA on the object vertices.
\\
We train a neural network $\net^a$, that uses the point feature $\mat{F}_j$ (same as~\cref{sec:SMPL_fitting}) corresponding to each query point $\vect{p}_j$ and predicts the global orientation of the object $\vect{a}_j = \net^a(\mat{F}_j), \vect{a}_j \in \mathbb{R}^9$.
We find that orientation prediction is unreliable if the query point is far from the object surface, hence we filter out points whose unsigned distance from the object surface $u^o_j$, is greater than a threshold $\epsilon=2cm$. The global orientation of the object is obtained by averaging the orientation predictions from the filtered points, $\vect{a}^o = \frac{1}{M}\sum_{j=1}^M \vect{a}_j$ where $M$ is the number of filtered points. Next, we compute the relative rotation between the current object orientation $\bar{\vect{a}}$ and the predicted object orientation $\vect{a}^o$, and use this to initialise the object rotation $\mat{R}^o = \vect{a}^o(\bar{\vect{a}}^T \bar{\vect{a}})^{-1}\bar{\vect{a}}^T$. We further run SVD on $\mat{R}^o$ and only keep the rotation matrix.
\\
Initialising $\mat{R}^o$ with the network predicted object orientation is crucial to avoid local minima during object fitting, as can be seen in \cref{fig:importance_of_initialization} and \cref{tab:ablation}.

\subsection{Refining human \& object models using contacts}
\begin{figure}[t!]
    \centering
    \begin{subfigure}{0.393\linewidth}
      \centering
      \caption{Input PC}
      \includegraphics[width=\linewidth]{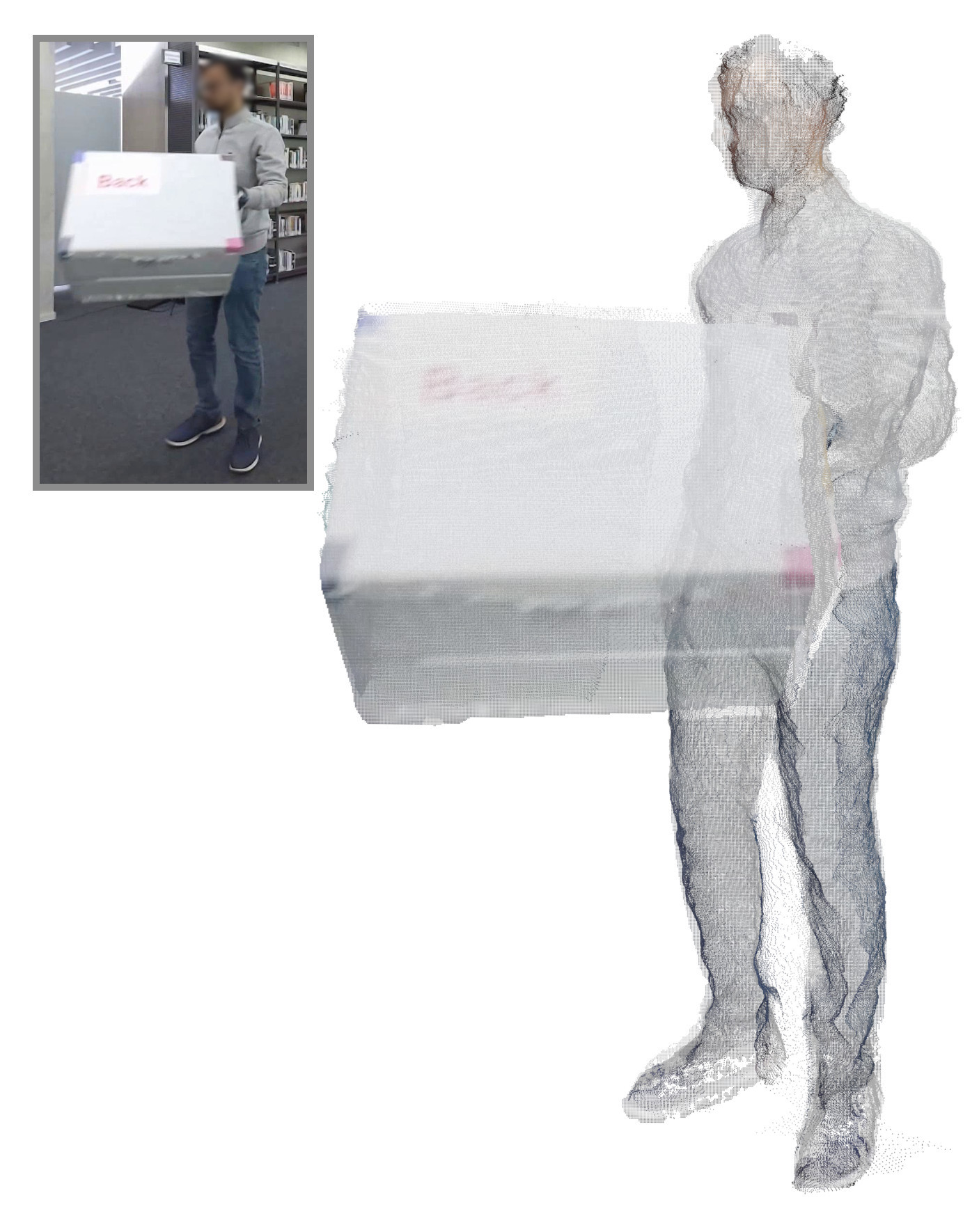}
    \end{subfigure}\hfill%
    \begin{subfigure}{\figFIVEsize\linewidth}
      \centering
      \caption{Ours w/o contacts}
      \includegraphics[width=\linewidth]{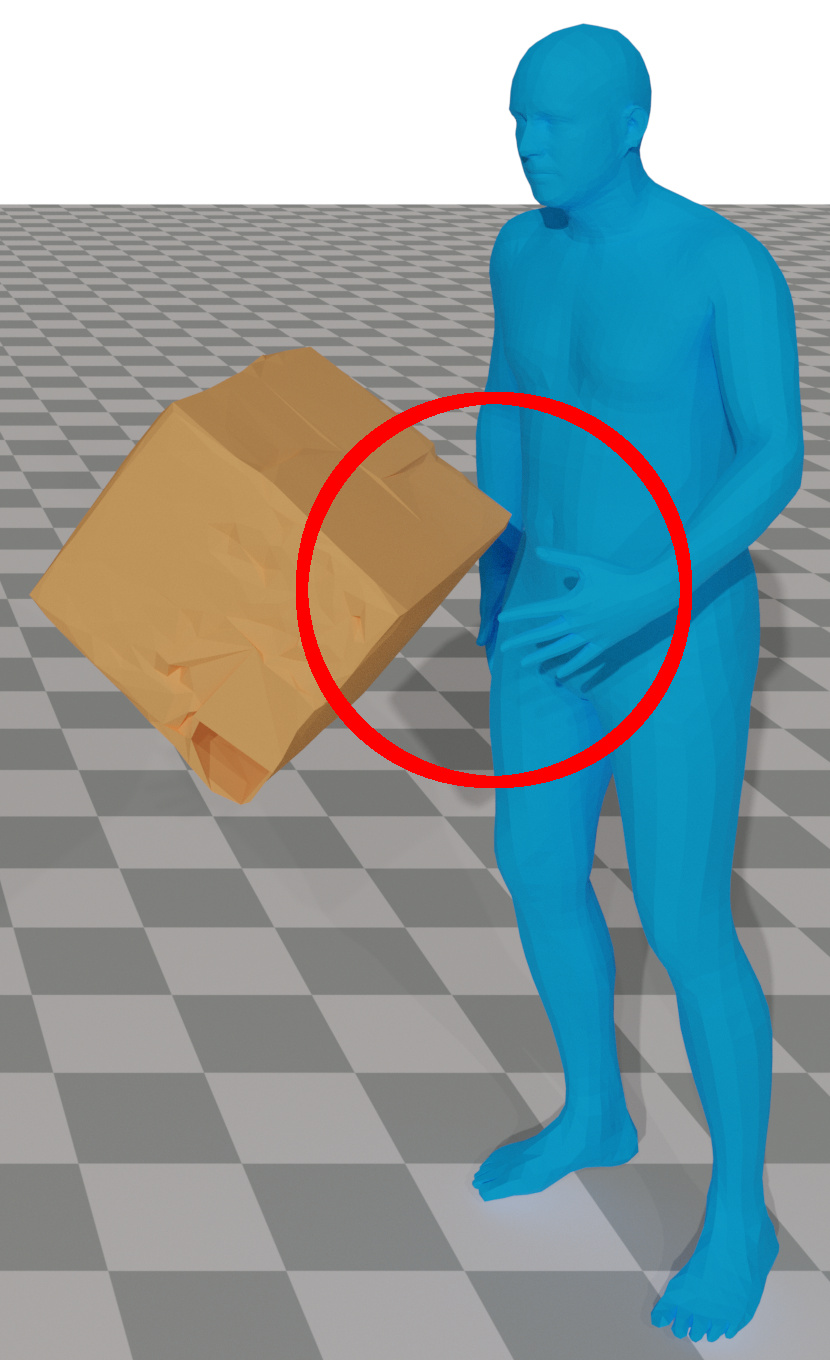}
    \end{subfigure}\hfill%
    \begin{subfigure}{\figFIVEsize\linewidth}
      \centering
      \caption{Ours}
      \includegraphics[width=\linewidth]{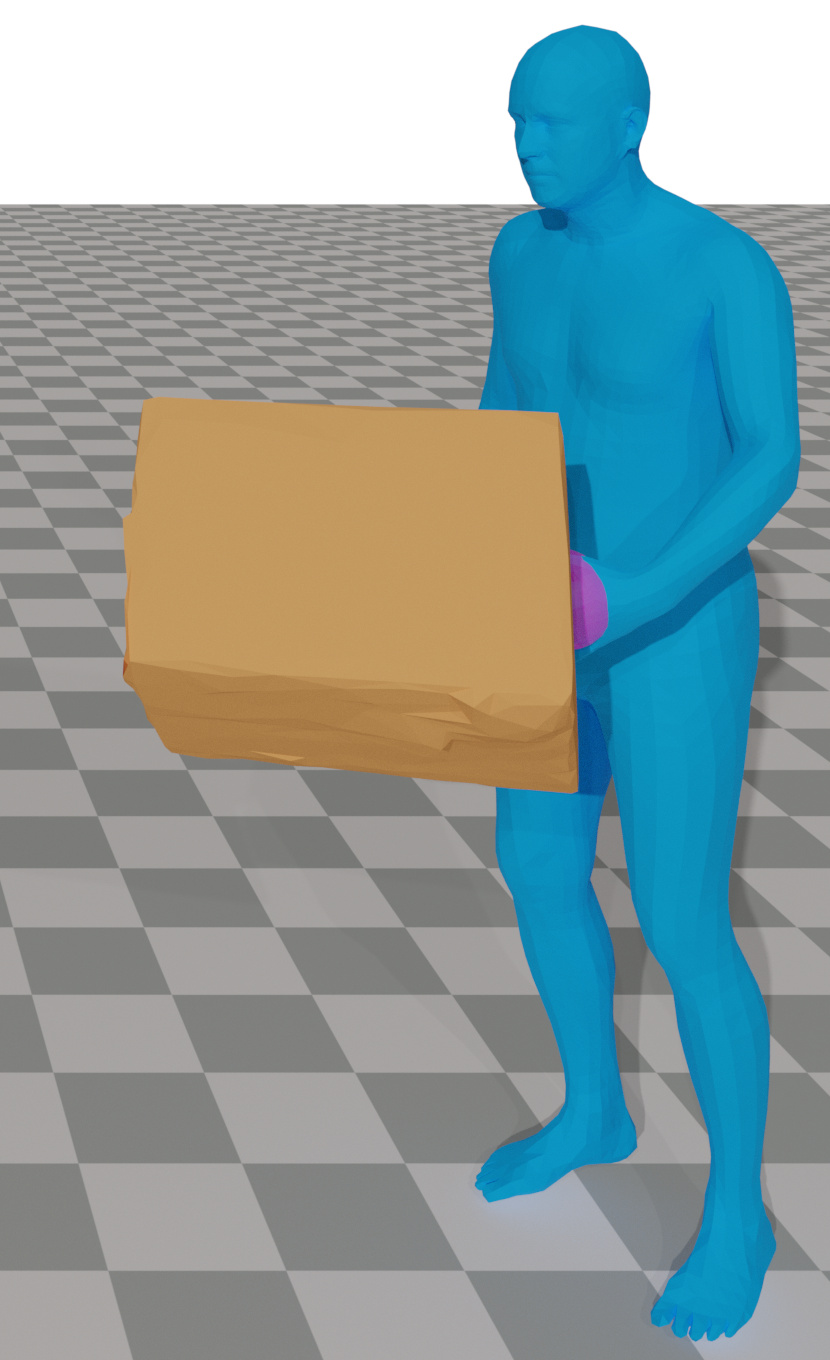}  
    \end{subfigure}
\caption{Without our network predicted contacts we observe artefacts like floating objects, leading to unrealistic tracking.}
\label{fig:importance_contacts}
\end{figure}

Our formulation above gives reasonably good human and object fits but does not ensure that human and object meshes satisfy the contacts predicted by the network. This often leads to floating objects and hovering hands see~\cref{fig:importance_contacts}, as human and object models are not in contact.
In this section we explicitly optimize the human and object meshes to fit the contacts predicted by the network.
We model contacts as vertices in the registered object mesh $\vect{v}_j^o \in \mat{W}^o$, that are very close to the input human $u^h_j<\epsilon$ and object $u_j^o<\epsilon$ surface.
Similarly to \cref{sec:object_fitting}, we use $\net^\text{udf}$ to obtain the unsigned distances $u^o_j, u^h_j$ and $\net^\text{corr}$ to obtain the correspondences $\vect{c}_j$ of these points to the SMPL model, respectively.
In order to filter query points close to human and object surfaces we compute a binary indicator matrix $\mathbbm{1}^c \in \mathbb{R}^N$ such that $\mathbbm{1}^c_j = 1$ iff $u^o_j<\epsilon, u^h_j<\epsilon$.
\begin{equation}
\label{eq:contact_optimization}
    E^\text{contact} = \sum_{j=1}^N \mathbbm{1}^c_j | \vect{v}^o_j - \smpl(\vect{c}_j, \pose, \shape) |_2.
\end{equation}
$E^\text{contact}$ allows us to jointly optimise the SMPL model and the object parameters $\mat{R}^o, \vect{t}^o$ to satisfy the contacts predicted by the network.

\subsection{Network training}
\noindent In this section we elaborate on training our networks.

\paragraph{Feature encoding.} We use a 3D CNN similar to IF-Net~\cite{ih_chibane2020} to obtain a voxel aligned multi-scale grid of features $\mat{F} = \net^\text{enc}(\humanpc, \objectpc)$.

\paragraph{Unsigned distance prediction.} To train the network $\net^\text{udf}$, we sample $N$ query points $\{\vect{p}_1, \dots, \vect{p}_N\}$ in 3D. For each query point $\vect{p}_j$ we obtain its point feature $\mat{F}_j$ (\cref{sec:SMPL_fitting}) and use this to predict the unsigned distance~\cite{chibane2020ndf} to human and object surface $u^o_j, u^h_j = \net^\text{udf}(\mat{F}_j)$.
\\
We jointly train $\net^\text{enc}, \net^\text{udf}$ with standard L2 loss. The GT for $u^o_j, u^h_j$ is easily available as our dataset contains GT SMPL and object fits allowing us to obtain GT distance of point $\vect{p}_j$ from the SMPL and object mesh.

\paragraph{SMPL correspondence prediction.} To train $\net^\text{corr}$, we use the point feature $\mat{F}_j$ of sampled query point $\vect{p}_j$ to predict its correspondence to the SMPL model $\vect{c}_j=\net^\text{corr}(\mat{F}_j)$.
\\
We jointly train $\net^\text{enc}, \net^\text{corr}$ using a standard L2 loss. Since we have the GT SMPL fit in our dataset we simply find the closest SMPL surface point for the query point $\vect{p}_j$ and use this as the GT correspondence.

\paragraph{Object orientation prediction.} To train the network $\net^a$, we use the point feature $\mat{F}_j$ of a sampled query point $\vect{p}_j$ to predict the global object orientation $\vect{a}_j=\net^a(\mat{F}_j)$.
\\
We jointly train $\net^\text{enc}, \net^a$ with standard L2 loss. We find that points far away from the object surface are unreliable in predicting the object orientation. Hence we only apply this loss to points that are close to the object, i.e. the GT $u^o_j<\epsilon$.
Since we have the GT object fit, we obtain the GT orientation by running PCA on the object mesh vertices and use the 3 principal axes in $\mathbb{R}^9$.

\section{Experiments}
\label{sec:experiments}
\begin{figure*}[t!]
    \centering
    \begin{subfigure}{\figSIXsize\linewidth}
      \centering
      \caption{Input PC}
      \includegraphics[width=\linewidth]{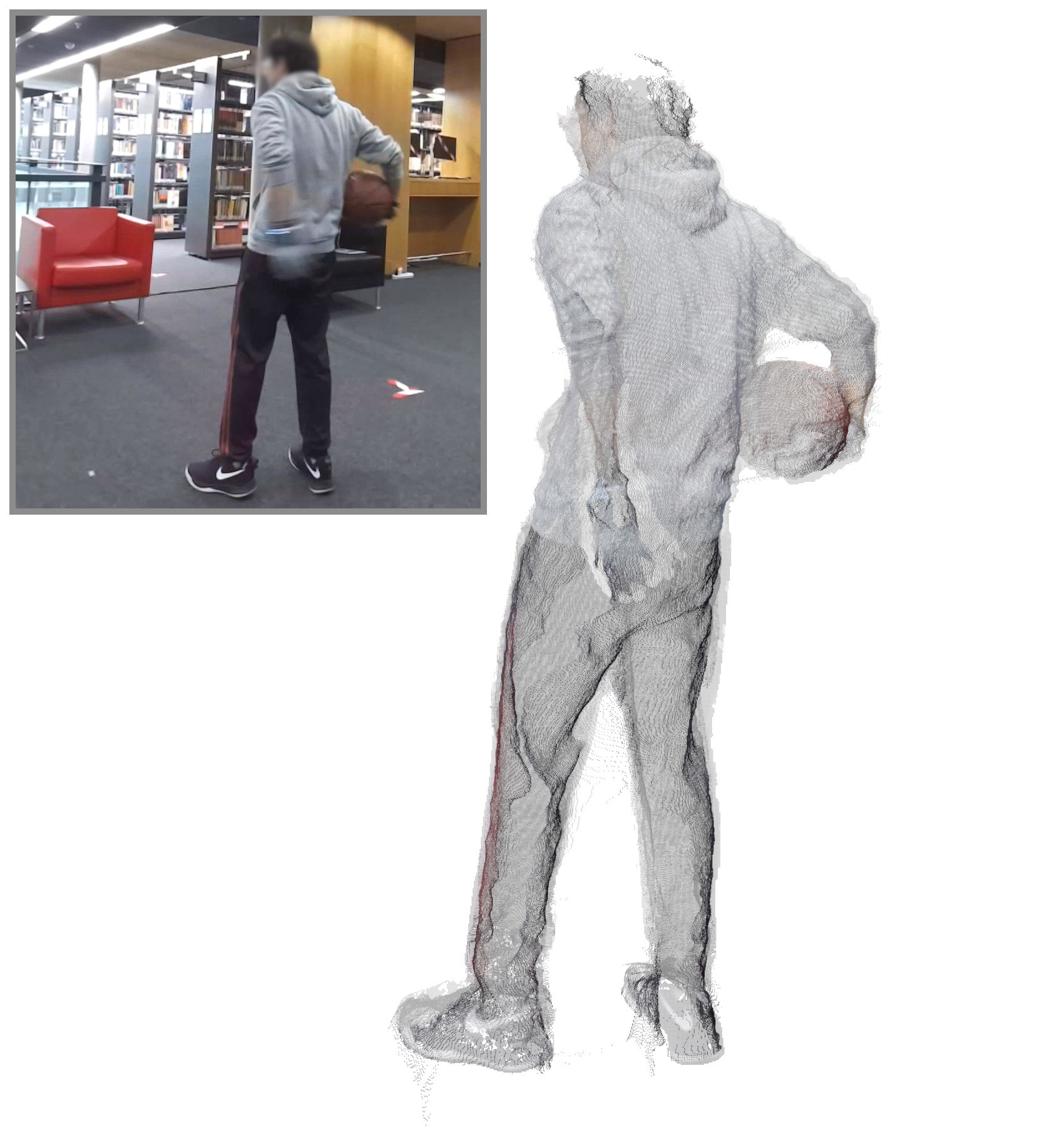}
    \end{subfigure}\hfill%
    \begin{subfigure}{\figSIXsize\linewidth}
      \centering
      \caption{PHOSA}
      \includegraphics[width=\linewidth]{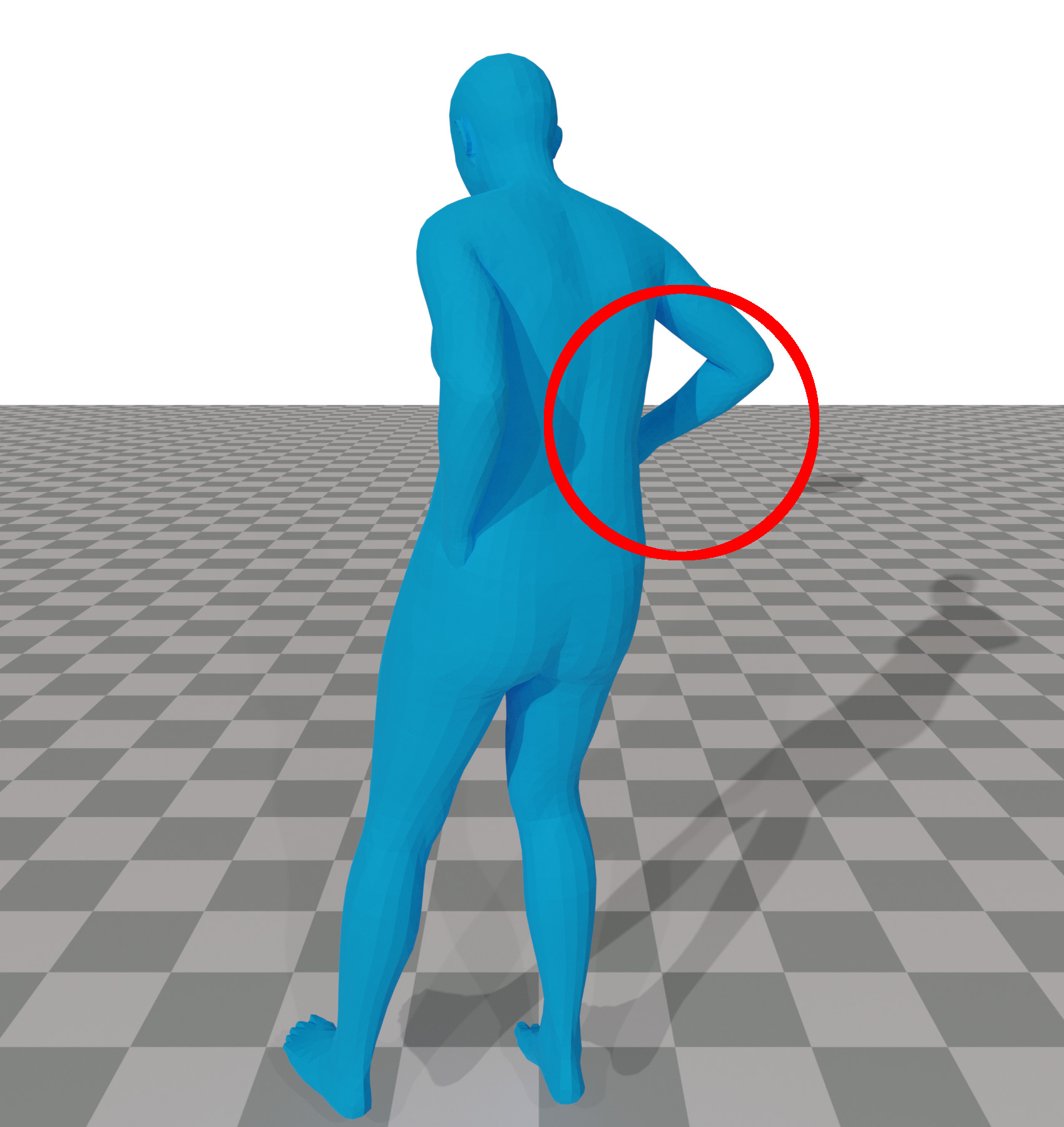}
    \end{subfigure}\hfill%
    \begin{subfigure}{\figSIXsize\linewidth}
      \centering
      \caption{Ours}
      \includegraphics[width=\linewidth]{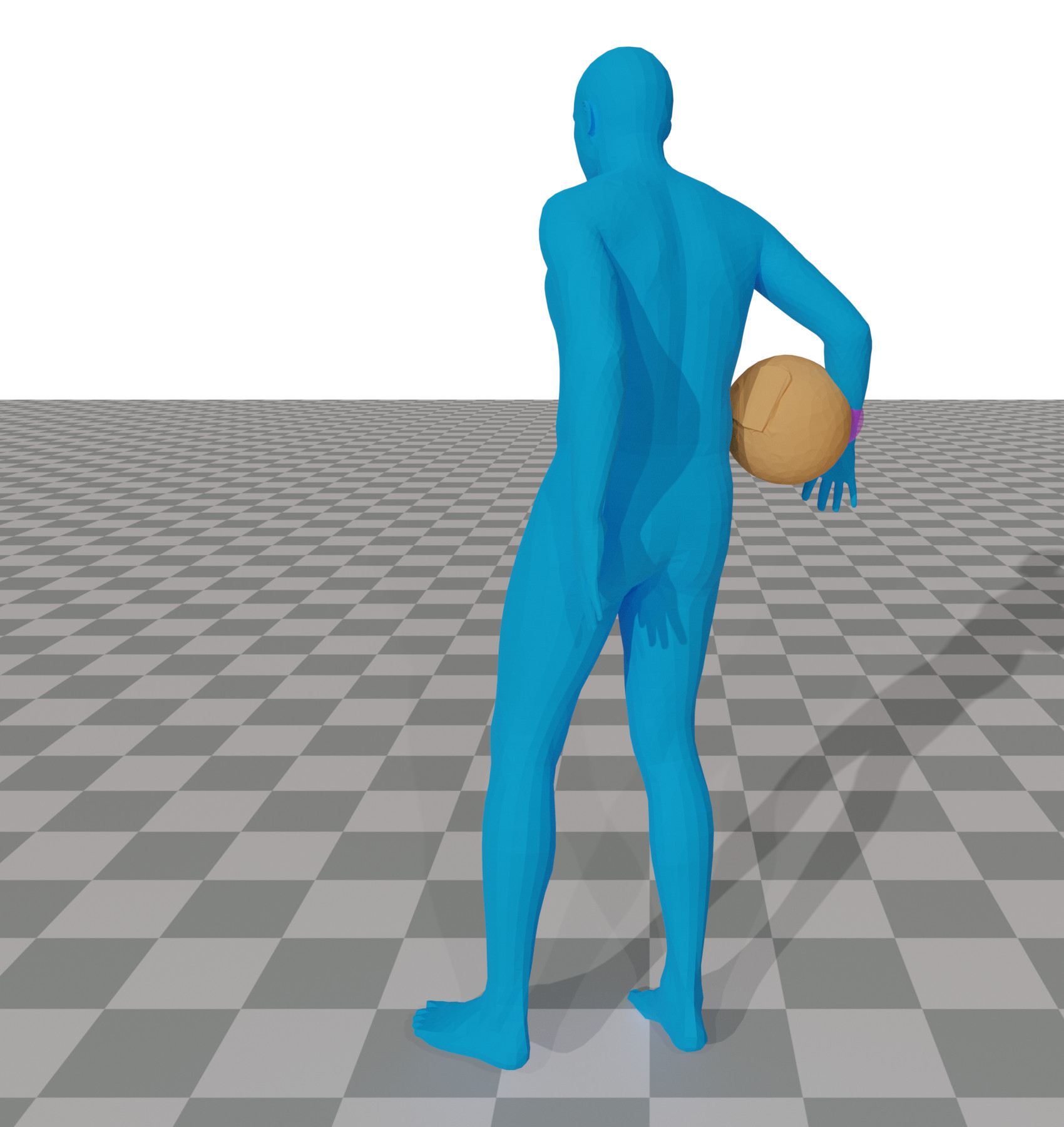}
    \end{subfigure}\hfill%
    \begin{subfigure}{\figSIXsize\linewidth}
      \centering
      \caption{PHOSA - side}
      \includegraphics[width=\linewidth]{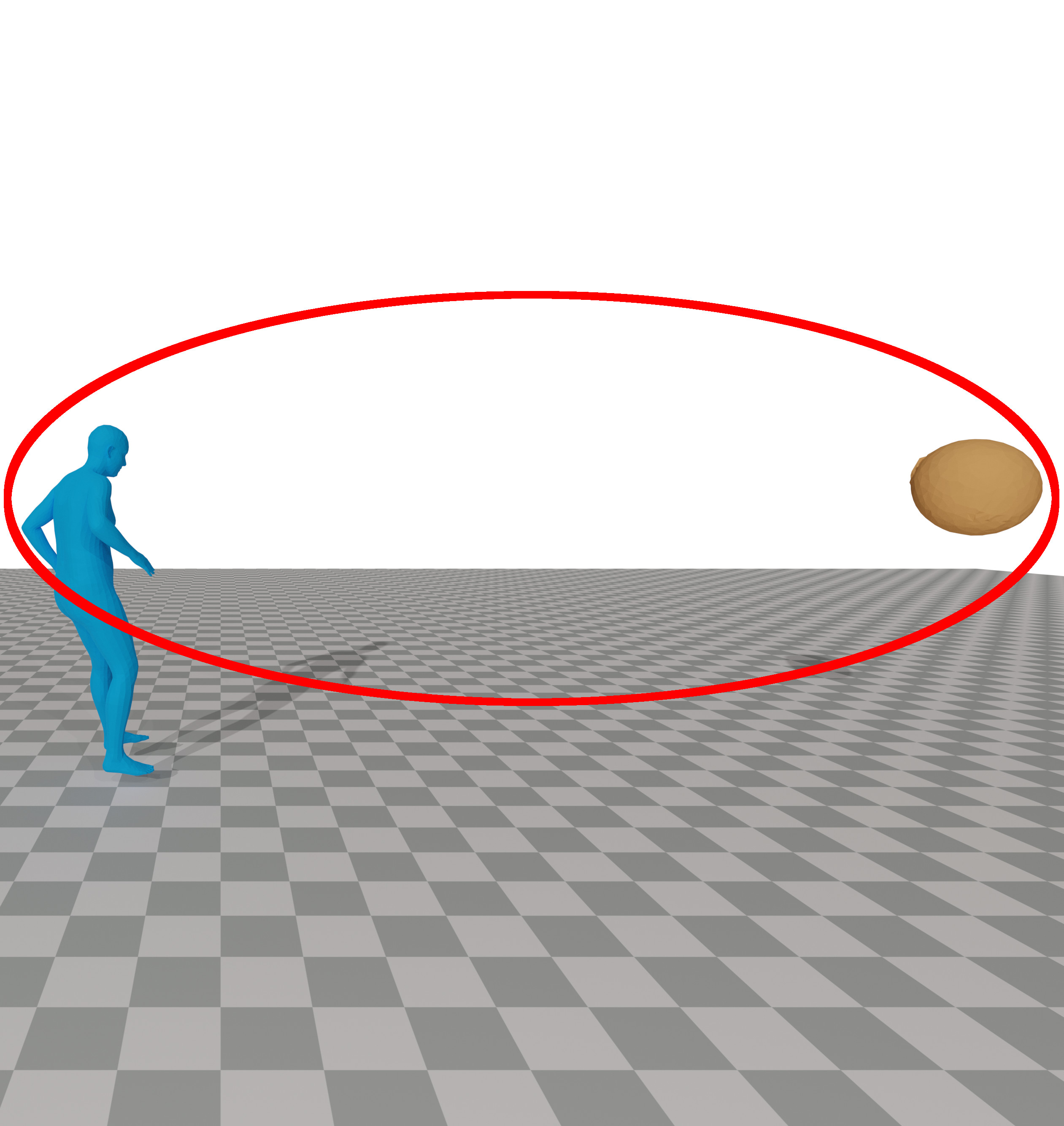}
    \end{subfigure}\hfill%
    \begin{subfigure}{\figSIXsize\linewidth}
      \centering
      \caption{Ours - side}
      \includegraphics[width=\linewidth]{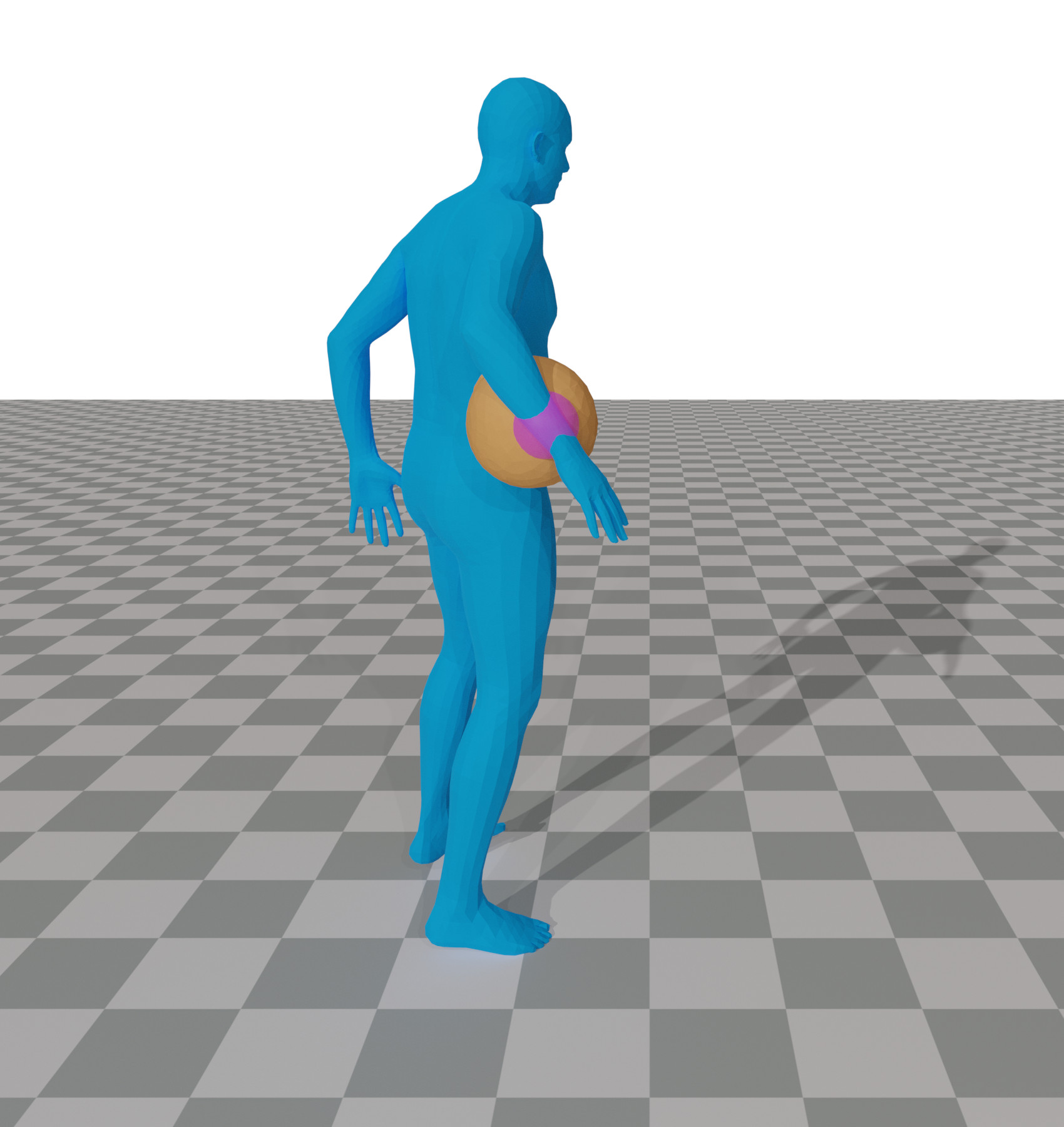}
    \end{subfigure}\hfill%
    \begin{subfigure}{\figSIXsize\linewidth}
      \centering
      \includegraphics[width=\linewidth]{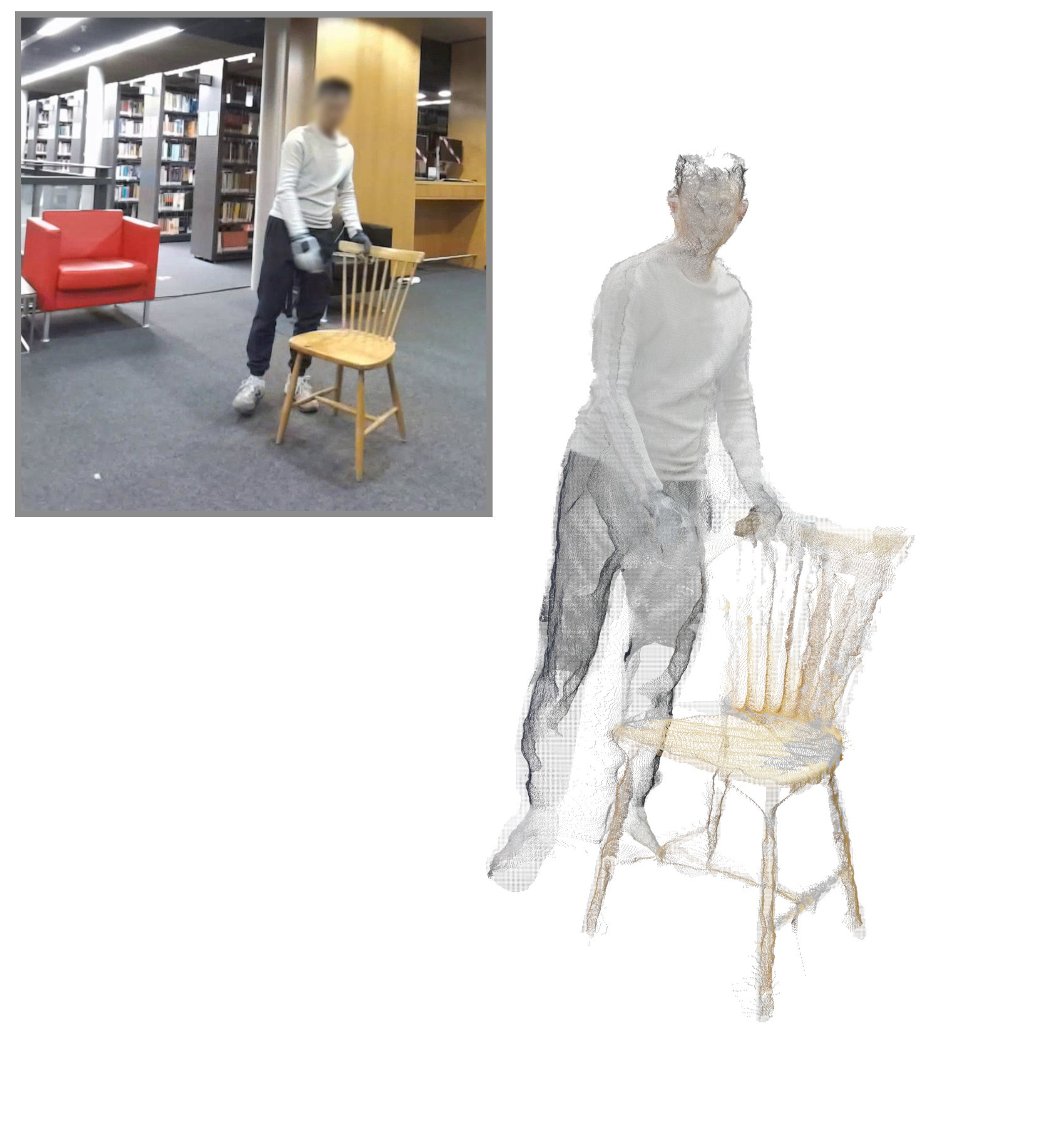}
    \end{subfigure}\hfill%
    \begin{subfigure}{\figSIXsize\linewidth}
      \centering
      \includegraphics[width=\linewidth]{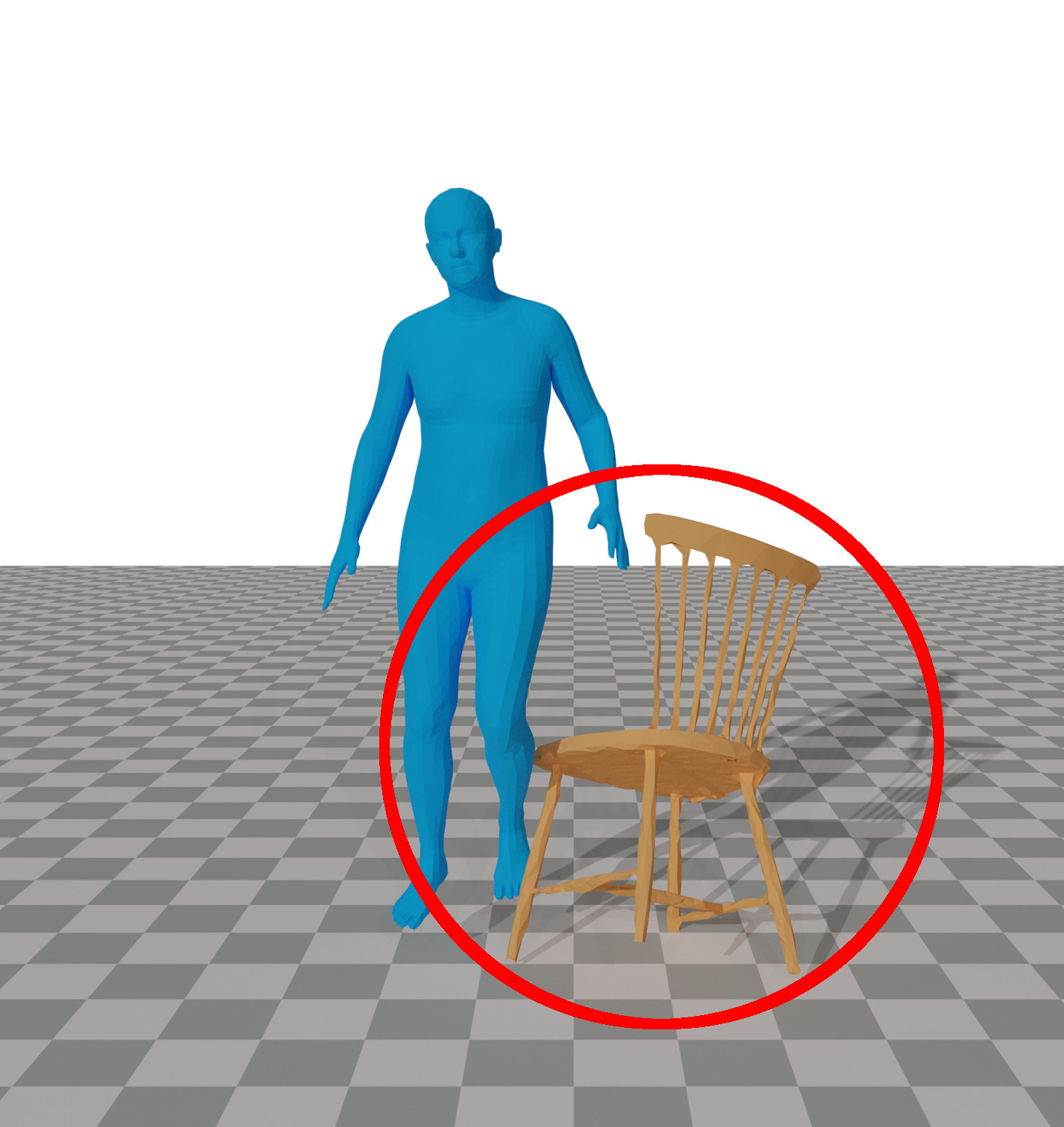}
    \end{subfigure}\hfill%
    \begin{subfigure}{\figSIXsize\linewidth}
      \centering
      \includegraphics[width=\linewidth]{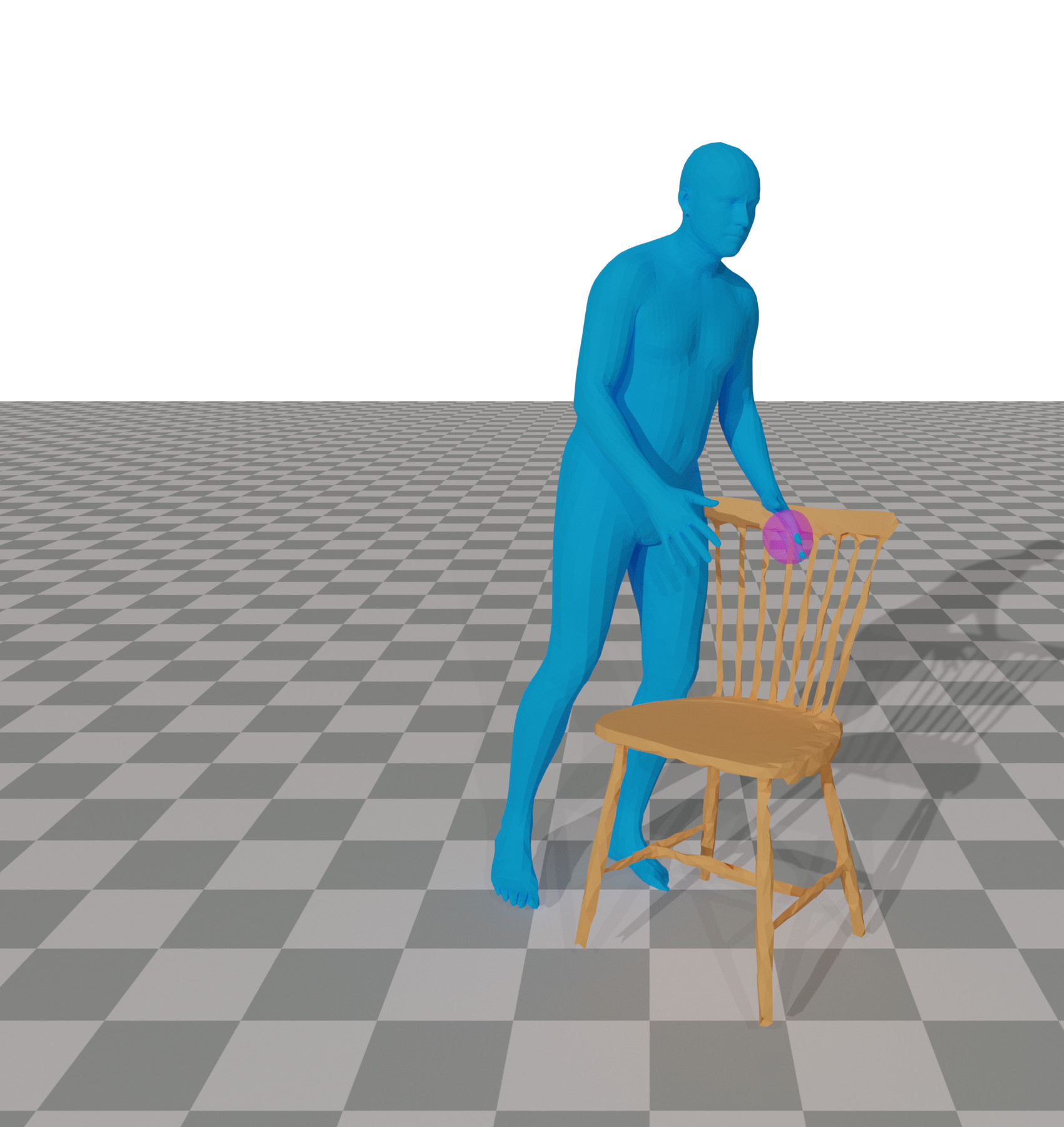}
    \end{subfigure}\hfill%
    \begin{subfigure}{\figSIXsize\linewidth}
      \centering
      \includegraphics[width=\linewidth]{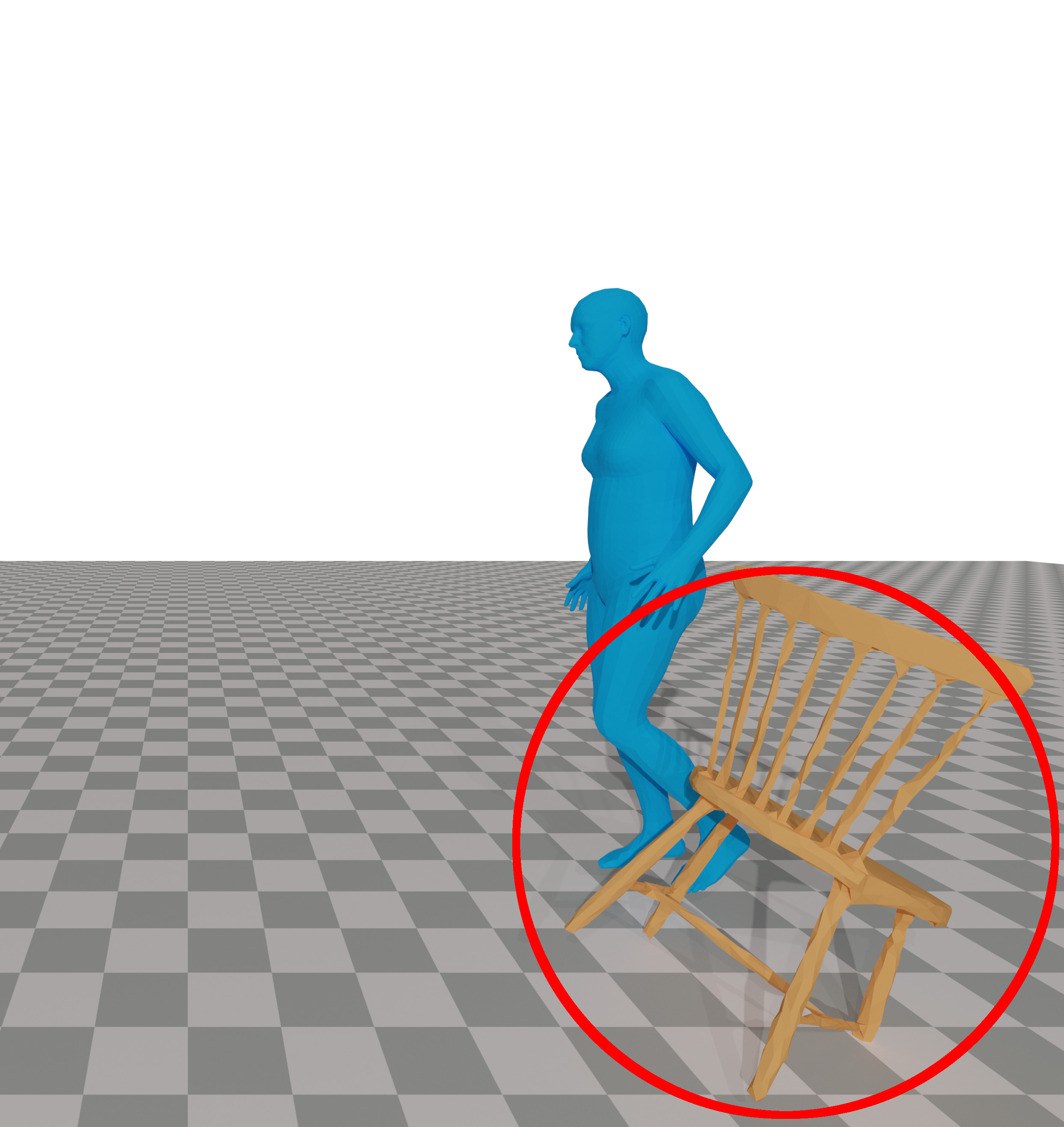}
    \end{subfigure}\hfill%
    \begin{subfigure}{\figSIXsize\linewidth}
      \centering
      \includegraphics[width=\linewidth]{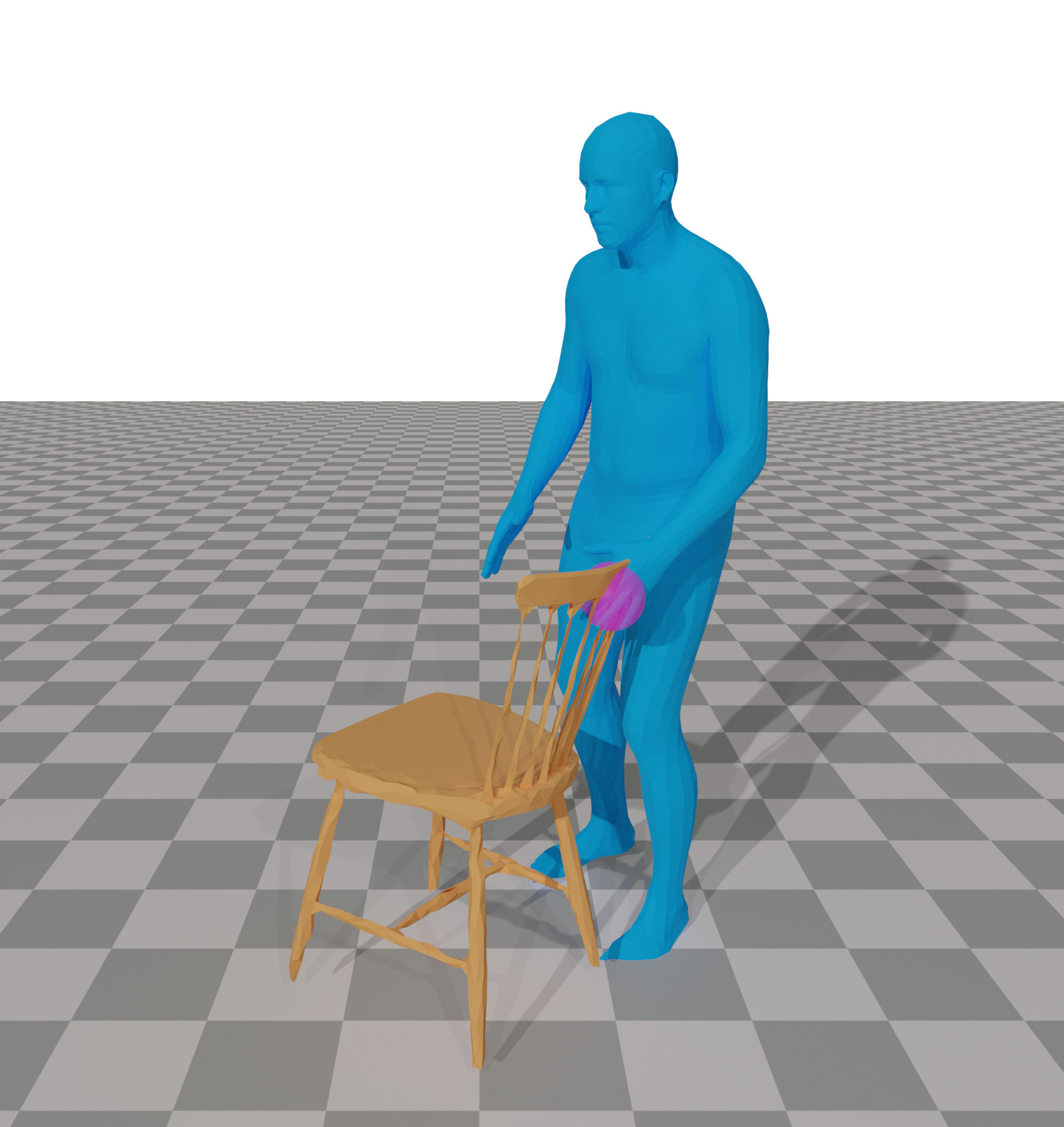}
    \end{subfigure}\hfill%
    \caption{We compare our method to track human, object and contacts with PHOSA~\cite{zhang2020phosa}. It can clearly be seen that our method can reason about the human-objects contacts and produces more accurate results.}
    \label{fig:tracking_results}
\end{figure*}

\begin{table}[t!]
  \centering
  \begin{tabular}{@{}lcc@{}}
    \toprule[1.5pt]
    \specialcell{\bf Method} & \specialcell{\bf SMPL v2v (cm)} & \specialcell{\bf obj. v2v (cm)}\\
    \midrule
    IP-Net~\cite{bhatnagar2020ipnet} & 6.61 & NA\\
    LoopReg~\cite{bhatnagar2020loopreg} & 9.12 & NA\\
    Fit to input & 16.15 & 26.09 \\
    PHOSA~\cite{zhang2020phosa} & 13.73 & 34.73 \\
    \midrule
    Ours & \textbf{4.99} & \textbf{21.20}\\
    \bottomrule[1.5pt]
  \end{tabular}
  \caption{We compare our method to obtain SMPL and object fits with IP-Net, LoopReg and PHOSA. We also show that directly fitting SMPL and object meshes to the input leads to sub-optimal performance.
  Our method not only obtains better fits, but unlike LoopReg and IP-Net, we can also fit the object.}
  \label{tab:comparison_IPNet}
\end{table}

In this section we compare our approach with existing methods. Our experiments show that we clearly outperform existing baselines. Next, we ablate our design choices and highlight the importance of contact and object orientation prediction in capturing human-object interactions.

\subsection{Comparing with PHOSA}
We find PHOSA~\cite{zhang2020phosa}, a method to reconstruct humans and objects from a single image, quite relevant to our work.
Although PHOSA uses only a single image whereas we use multi-view images, thus giving our method an advantage, it is still the closest competing method. We run Procrustes alignment on PHOSA results to remove depth ambiguity.
\\
It should be noted that PHOSA depends on pre-defined fixed contact regions whereas our approach can freely predict full-body contacts and PHOSA uses hand crafted heuristics to model contacts whereas our approach learns contact modelling from data, making our method more scalable.
We compare our method with PHOSA in~\cref{fig:tracking_results} and \cref{tab:comparison_IPNet}, and clearly outperform it.

\subsection{Why not fit human and object models directly to point clouds?}
Since there are no existing methods that can jointly track humans, objects and the contacts from a multi-view input, we create an obvious baseline where we fit the SMPL and object meshes directly to the input point cloud. We show (\cref{tab:comparison_IPNet}) that direct fitting easily gets stuck in local minima. This is because the point clouds are very noisy and large parts are missing due to heavy occlusion between the person and the object during interactions.
Our network, on the other hand, can implicitly reason about missing parts, thus generating more accurate results.

\begin{table}[t!]
  \centering
  \begin{tabular}{@{}lcc@{}}
    \toprule[1.5pt]
    \specialcell{\bf Method} & \specialcell{\bf SMPL v2v (cm)} & \specialcell{\bf obj. v2v (cm)}\\
    \midrule
    A) Ours w/o ori. & 4.98 & 24.02\\
    B) Ours w/o cont. & \textbf{4.96} & 21.28\\
    \midrule
    C) Ours & 4.99 & \textbf{21.20}\\
    \bottomrule[1.5pt]
  \end{tabular}
  \caption{We analyse the importance of (A) object orientation prediction and (B) contact prediction for our method. It can be seen that object orientation prediction noticeably improves object localisation error. The effect of contact loss is not significant quantitatively but makes noticeable difference qualitatively see~\cref{fig:importance_contacts}.}
  \label{tab:ablation}
\end{table}
\subsection{Why can't existing human registration approaches be extended to our setting?}
There are no direct baselines that can jointly track humans, objects, and contacts from multi-view input. There are works~\cite{bhatnagar2020ipnet, bhatnagar2020loopreg} that pursue similar ideas of predicting correspondences and fitting SMPL to the human point cloud.
In this subsection we explore their suitability in our setting.

\paragraph{Comparison to IPNet~\cite{bhatnagar2020ipnet}} 
IPNet takes as input a human point cloud and predicts an implicit reconstruction of the human and sparse correspondences to the SMPL model, which enables its fitting to the implicit reconstruction.
\\
This approach has three major disadvantages. First, querying occupancies for a $128^3$ grid to obtain implicit reconstruction is expensive. Second,
it predicts occupancies which requires water-tight surfaces. And third, running traditional Marching Cubes makes occupancy prediction non-differentiable w.r.t. SMPL fitting.
\\
Our formulation in~\cref{eq:SMPL_fitting,eq:SMPL_fitting_correspondences} alleviates these problem as we can fit SMPL by only querying $N=30k$ points instead of $128^3 (\sim 2M)$ points. Since we use unsigned distance prediction, our method can work with non-water tight surfaces. We can also fit SMPL directly to unsigned distance predictions, thus removing the requirement for Marching Cubes.
We compare our approach with IP-Net~\cite{bhatnagar2020ipnet} (trained on our dataset) in ~\cref{tab:comparison_IPNet} and show that we obtain better performance than IP-Net at much lower cost ($30k$ (ours) vs. $\sim2M$ (IP-Net) query points and no Marching Cubes). This shows that our formulation is superior than IPNet even for human registration. 
We can additionally handle objects and interactions.
Qualitative comparisons are given in the supplementary material.

\paragraph{Comparison with LoopReg~\cite{bhatnagar2020loopreg}}
LoopReg fits SMPL to the input point cloud by explicitly predicting correspondences. We find the idea interesting and use their diffused SMPL formulation in our method. LoopReg is, however, not directly applicable in our setting as it assumes a noise free and complete human point cloud. When the point cloud is incomplete due to occlusions, no correspondences are predicted for missing parts. Since LoopReg can only use surface points for fitting, this makes registration inaccurate.
\method{} handles this case by using distances to the SMPL surface(~\cref{eq:SMPL_fitting_correspondences}) predicted for each of the sampled query points to fit the body model, thus allowing the use of non-surface points for fitting.
This is important as the Kinect point cloud is noisy.
We outperform LoopReg~\cite{bhatnagar2020loopreg} (trained on our dataset) and show (\cref{tab:comparison_IPNet}) that our formulation is robust to missing parts and noisy input.

\subsection{Importance of contacts}
\label{sec:importance_contacts}
In this experiment we show that our network predicted contacts are key for physically plausible tracking. Even though quantitative difference is not significant~(\cref{tab:ablation}), it can be seen in~\cref{fig:importance_contacts} that without contact information, the human and the object do not lock into the correct location. Hence, we notice unnatural results like floating objects. Using our contact prediction alleviates such issues.

\noindent We encourage the readers to see our supplementary document for detailed discussion regarding limitations and future work with \method{}.

\section{Conclusions}
\label{sec:conclusions}
We have presented \method{}, the first methodology to jointly track humans, objects and explicit contacts in natural environments.
By introducing neural networks to predict correspondences to a 3D human body model along with unsigned distance fields defined over human and object surfaces, we are able to accurately model human-object contacts.
We further integrate such neural predictions into a proposed joint registration method resulting in the robust 3D tracking of human-object interactions.
\\
Along with our proposed method we also provide \method{}, the \emph{largest} dataset of RGBD sequences and annotated humans, objects, and contacts to date.
\method{} dataset is the \emph{first} benchmark for the part of the research community interested in modelling human-object interactions. We propose real-world challenges like reconstructing humans and object from a single RGB image, tracking human-object interactions
from multiple and single-view RGB(D) input, pose estimation etc.
Our dataset together with our code is released in order to stimulate future research in this important emerging domain.

{\small
\paragraph{Acknowledgements}
Special thanks to RVH team members \cite{rvh_grp}, and reviewers, their feedback helped improve the manuscript.
This work is funded by the Deutsche Forschungsgemeinschaft (DFG, German Research Foundation) - 409792180 (Emmy Noether Programme,
project: Real Virtual Humans), German Federal Ministry of Education and Research (BMBF): Tübingen AI Center, FKZ: 01IS18039A and ERC Consolidator Grant
4DRepLy (770784). Gerard Pons-Moll is a member of the Machine Learning Cluster of Excellence, EXC number 2064/1 – Project number 390727645.
}

{\small
\bibliographystyle{ieee_fullname}
\bibliography{egbib}
}

\end{document}